\documentclass[letterpaper,journal]{IEEEtran}

\newcommand{\vs}{\emph{vs.}}

\usepackage{amsmath,amsfonts}
\usepackage{graphicx}
\usepackage{booktabs}
\usepackage{multirow}
\usepackage[table,xcdraw]{xcolor}
\usepackage[normalem]{ulem}   
\usepackage{url}
\usepackage{cite}
\usepackage{hyperref}
\hypersetup{colorlinks=false}

\begin{document}

\title{Boosting Robustness for All-Weather Self-Supervised \\Depth Estimation in Autonomous Driving}

\author{%
  Mengshi~Qi, \IEEEmembership{Member, IEEE},%
  ~Xiaoyang~Bi, Xianlin~Zhang, and Huadong~Ma, \IEEEmembership{Fellow, IEEE}%
  \thanks{This work is partly supported by the Funds for the NSFC Project (Grant 62572072) and Beijing Natural Science Foundation (L243027). (\emph{Corresponding author: Mengshi Qi and Xianlin Zhang~(email:~qms@bupt.edu.cn)})}
\thanks{M. Qi, X. Bi, X. Zhang and H. Ma are with State Key Laboratory of Networking and Switching Technology, Beijing University of Posts and Telecommunications, China.}
}

\maketitle

\begin{abstract}
Self-supervised depth estimation is challenging for safe autonomous driving under various adverse weather conditions due to sensor perception degradation. These challenges arise from two main aspects. Firstly, adverse conditions can distort pixel correspondences and violate the assumptions embedded in the self-supervised loss function, leading to erroneous depth predictions. Secondly, while radar is a widely adopted sensor in adverse weather conditions, the sparse distribution of radar points in the Point of View (POV) poses challenges for self-supervised fusion. To address these issues, we introduce a novel self-training pipeline using unpaired real all-weather data through multi-teacher distillation and robust radar fusion. We propose the Uncertainty-Aware Multi-Teacher Distillation method to generate diverse teacher models with different adverse condition inputs, and then employ uncertainty modeling to weigh the knowledge distillation loss. Additionally, we design the POV-BEV Radar Fusion approach, which leverages camera-pixel ray constraints to establish connections between the camera's Point of View (POV) and the radar's Bird’s-Eye View (BEV). This approach enables the utilization of denser radar points, effectively capturing the complementary perspectives of both POV and BEV. Extensive quantitative and qualitative experiments demonstrate the robustness of our proposed method on all-weather datasets, achieving state-of-the-art performance.  Our code and models are available
at \text{https://github.com/MICLAB-BUPT/RobustDepth}.
\end{abstract}


\section{Introduction}
\label{sec:intro}

In the field of autonomous driving~\cite{ye2025safedriverag, lv2025t2sg, liao2026improving}, depth estimation is crucial for understanding 3D environments~\cite{almalioglu2022deep, deng2026active}. Monocular depth estimation aims to regress the depth of each pixel in an image, where supervised methods learn from Lidar data, and self-supervised methods estimate depth through the perspective differences between two adjacent frames. Current popular perception algorithms, such as 3D detection and planning, require the establishment of a BEV feature space~\cite{philion2020lift}, which in turn necessitates accurate depth estimation. However, using sensors such as Lidar to generate large-scale ground truth (GT) labeled data for training depth estimation networks is expensive, which limits its application scenarios~\cite{godard2019monodepth2}. Hence self-supervised depth estimation methods have garnered significant attention due to their ability to be trained on large-scale datasets without requiring labeled data~\cite{zhou2017unsupervised,woo2025prodepth,zhao2022monovit, godard2019monodepth2,wang2023planedepth,yan2021channel,feng2022disentangling}.

\begin{figure}[!t]
\begin{center}
\includegraphics[width=1.00\linewidth]{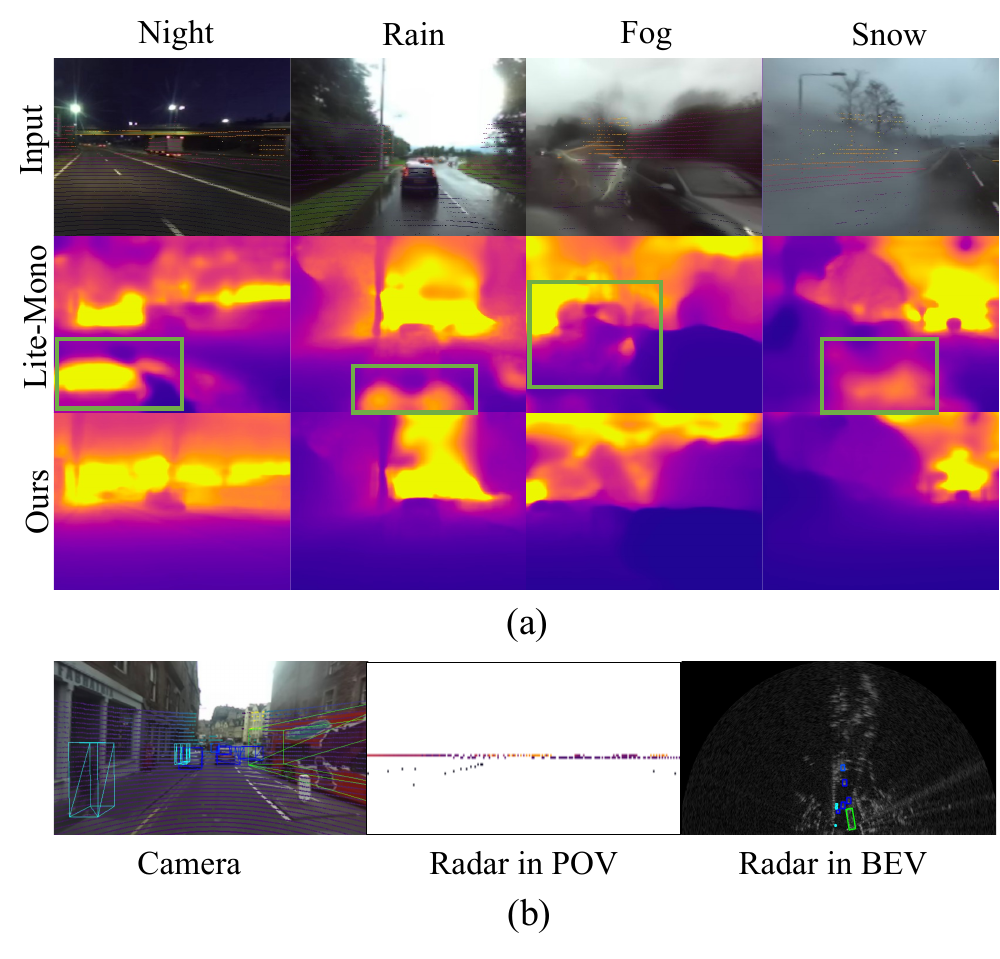}
\vspace{-1.0cm}
\end{center}
   \caption{(a) The state-of-the-art self-supervised depth estimation model, Lite-Mono~\cite{zhang2023lite}, predicts errors on real all-weather data (see Row 2), such as misestimating the ground as a hole at nighttime. While our proposed method can address these issues, enhancing robustness (see Row 3). (b) The radar in the Point of View (POV) perspective has minimal height, while in the Bird's Eye View (BEV), it shows a denser distribution of points. Integrating radar data from the BEV perspective can be used to improve depth prediction (Box annotations are used for visualization).}
\label{fig:intro}
\vspace{-0.2cm}
\end{figure}

The diversity and complexity of driving conditions pose a major challenge for existing deep learning-based monocular depth estimation algorithms. Low night illumination and adverse weather conditions, such as raindrops, fog, and snow, significantly degrade their performance. These phenomena break the assumptions embedded in the self-supervised loss function, particularly the brightness-consistency assumption, which states that corresponding pixels in two adjacent frames should have the same color. As shown in the Fig.~\ref{fig:intro}~(a):
1)~Due to the effect of rain, the ground becomes texture-less, and the depth prediction network then cannot correctly construct the correspondence between adjacent frames.
2)~The varying lights of traffic signals at night do not conform to the constant illumination assumption and are incorrectly recognized as objects.
3)~On foggy days, the refraction and blur caused by raindrop occlusion in front of the camera and snowflake occlusion in snowy weather make the changes in lighting unpredictable.
4)~The ground at night appears to flicker due to vehicle headlights, resulting in a hole in the depth prediction in this region.
The above issues pose significant challenges for self-supervised depth prediction~\cite{saunders2023self, gasperini2023robust}.

Recent research on self-supervised monocular depth estimation has increasingly focused on achieving robustness under all-weather conditions~\cite{lv_disent}. Directly using real all-weather data for training can lead to undesirable outputs~\cite{gasperini2023robust}. Most existing work aims to improve robustness by utilizing paired data from clear and adverse weather conditions~\cite{gasperini2023robust, saunders2023self, tosi2024diffusion, wang2024weatherdepth, liu2021self, wang2021regularizing, vankadari2023sun}. These approaches often rely on generative models or diffusion processes to synthesize adverse condition data. However, training such generative models is still a challenging task, highly dependent on the quality of training data, and model architecture, and is computationally expensive. On the other hand, the synthetic adverse weather data generated by these models still exhibits a domain gap~\cite{saunders2023self} compared to real-world adverse weather data. Furthermore, although large-scale all-weather datasets~\cite{sheeny2021radiate} have been released, training methods on real all-weather data remain underexplored. More importantly, fusing radar data can achieve more robust depth prediction. However, as shown in Figure~\ref{fig:intro}~(b), the radar points in Point of View (POV) are much sparser than in Bird’s-Eye View (BEV) and lack the complementary perspective provided by BEV, which offers a more global spatial awareness. This limitation restricts spatial perception and ultimately degrades performance.

To address the above-mentioned issues, we propose a novel self-supervised depth estimation method by adopting multi-teacher distillation to achieve effective training on all-weather datasets. Specifically, we train diverse and high-quality teacher models, each specialized in different weather conditions as ``weather experts'', while incorporating a novel distillation loss to enhance knowledge transfer. Then we introduce an uncertainty-aware distillation loss to guide the student model in selectively learning from the teacher models, as these teacher models are not always reliable. Furthermore, we propose a novel camera-radar cross-view fusion method to exploit the dense radar points in bird’s eye view to enhance robustness. 

Our main contributions can be summarized as follows:\\
\textbf{(1)} We introduce a novel self-supervised depth estimation method to generate multiple ``weather-expert'' teachers from all-weather data, and propose the uncertainty-aware multi-teacher distillation strategy that jointly assesses the pseudo-label quality of multiple teachers for selective knowledge transfer.\\ 
\textbf{(2)} We propose a novel camera-radar cross-view fusion method that leverages camera-pixel ray constraints to establish cross-attention between the camera's Point of View (POV) and the radar's Bird’s-Eye View (BEV). Additionally, it incorporates radar’s POV to fully utilize the complementary perspectives of radar for enhancement. \\ 
\textbf{(3)} We validate the effectiveness of our method from both quantitative and qualitative perspectives on real and complex all-weather datasets, achieving a 26\% absRel error reduction on the RADIATE dataset and a 23\% reduction in night conditions on the nuScenes dataset compared to state-of-the-art methods.
\section{Related Work}
\label{sec:related_work}

\textbf{Monocular Depth Estimation.}~Early research approached depth prediction as a regression problem, exploring various deep learning architectures~\cite{dcsam, qi2025action, qi2020few}. Such as Eigen~\textit{et al.}~\cite{eigen2014depth} first used a CNN-based structure, while Laina~\textit{et al.}~\cite{laina2016deeper} later introduced a residual structure. Recently, some studies have treated supervised depth prediction as an ordinal regression problem, achieving significant progress~\cite{bhat2021adabins, lee2019big, fu2018deep}. Zhou~\textit{et al.}~\cite{zhou2017unsupervised} pioneered using view synthesis as a supervisory signal by estimating the depth of a single frame and the pose between consecutive frames, enabling end-to-end learning. Godard~\textit{et al.}~\cite{godard2019monodepth2} further advanced self-supervised methods with MonoDepth2. Several works address challenges like estimating moving objects~\cite{woo2025prodepth,sun2024dynamo} and the scale-ambiguous problem~\cite{zhang2022towards, bian2019unsupervised, wang2021can, zhu2023ec}. Subsequent studies introduced temporal network structures by constructing cost volumes~\cite{woo2025prodepth, zou2025m, wu2023self, miao2023ds}. Recently, researchers have explored Vision Transformer-like structures for depth estimation~\cite{zhang2023lite, zhao2022monovit}. To deal with dynamic objects that break the static-scene assumption, several methods design dedicated strategies for moving regions~\cite{sun2024dynamo, woo2025prodepth, qi2019sports, qi2021semantics, qi2026explainable}. A parallel line of work addresses the inherent scale ambiguity of monocular predictions through test-time refinement or additional geometric constraints~\cite{zhang2022towards, bian2019unsupervised, wang2021can, zhu2023ec}. Although these single- and multi-frame designs achieve strong results in daytime, they are seldom evaluated under adverse weather and tend to degrade sharply when trained or tested in rain, fog, snow, or at night. However, previous work has primarily focused on improving depth estimation in clear weather while overlooking robustness in adverse weather conditions. In contrast, we enhance the training pipeline to improve model robustness for all-weather depth estimation.

\noindent\textbf{Robustness in Adverse Conditions.}~Self-supervised monocular depth estimation struggles in adverse weather~\cite{gasperini2023robust, saunders2023self, tosi2024diffusion, wang2024weatherdepth, liu2021self, wang2021regularizing, vankadari2023sun}. Direct training with adverse weather data often leads to underfitting~\cite{gasperini2023robust}. Current methods use GANs or data augmentation to create paired clear and adverse condition data~\cite{gasperini2023robust, tosi2024diffusion}. Early robustness attempts often target a single condition in isolation, for example by translating daytime images into nighttime with GANs or by applying image-enhancement preprocessing~\cite{liu2021self, wang2021regularizing, vankadari2023sun}; although useful, they generalize poorly across the full spectrum of weather. More recent approaches instead synthesize large-scale paired adverse data using GANs or diffusion models and learn in a teacher-student manner~\cite{tosi2024diffusion, gasperini2023robust}. Nevertheless, the synthesized data inevitably carries a synthetic-to-real domain gap and frequently depends on extra annotations, which restricts its utility for unseen conditions. Such as Saunders~\textit{et al.}~\cite{saunders2023self} introduced pseudo pose and depth loss, Liu~\textit{et al.}~\cite{liu2021self} proposed domain-invariant learning, and Wang~\textit{et al.}~\cite{wang2024weatherdepth} used curriculum contrastive learning to prevent forgetting. Besides, distillation learning is adopted to generate pseudo labels from a pretrained teacher model to train a student model~\cite{gasperini2023robust, tosi2024diffusion}. Distillation has also been used to transfer a clear-weather teacher's knowledge so as to stabilize training under adverse conditions~\cite{gasperini2023robust, tosi2024diffusion, Guo2024UNIKD, wen2024class}, yet such methods rely on a single teacher and treat all pseudo-labels as equally trustworthy, limiting the diversity of transferred knowledge. Uncertainty-aware learning further improves reliability by reweighting unreliable pixels~\cite{poggi2020uncertainty, kendall2017uncertainties, kendall2018multi}. However, existing single-teacher uncertainty methods~\cite{poggi2020uncertainty, kendall2017uncertainties} treat each teacher independently or estimate the confidence of a single prediction. On the other hand, supervised methods focus on creating dense radar-to-pixel maps for fusion with camera data~\cite{long2021radar, lo2021depth, singh2023depth, li2023sparse} in a two-stage process with LiDAR data for training. In the field of self-supervised depth estimation, Gasperini~\textit{et al.} use radar as a weak supervisory signal for moving object estimation~\cite{gasperini2021r4dyn}, but this approach also requires additional box annotations and velocity data. Moreover, these approaches do not address the sparsity of radar data during fusion. Radar is particularly attractive for all-weather perception because its long wavelength is largely unaffected by rain, fog, or darkness, yet its extreme sparsity makes direct fusion non-trivial.  Furthermore, UAMTD is not a mere combination of off-the-shelf modules; we are the first to extend uncertainty estimation to the \emph{joint} assessment of multiple teachers' pseudo-label quality, arbitrating among multiple supervisory sources and accounting for inter-teacher interactions. Second, our POV-BEV Radar Fusion method fuses camera and radar data from both POV and BEV perspectives, leveraging the denser radar points in BEV space.

\section{Proposed Approach}

\subsection{Overview}

\begin{figure*}[!t]
  \centering
  \resizebox{1.\linewidth}{!}{\includegraphics{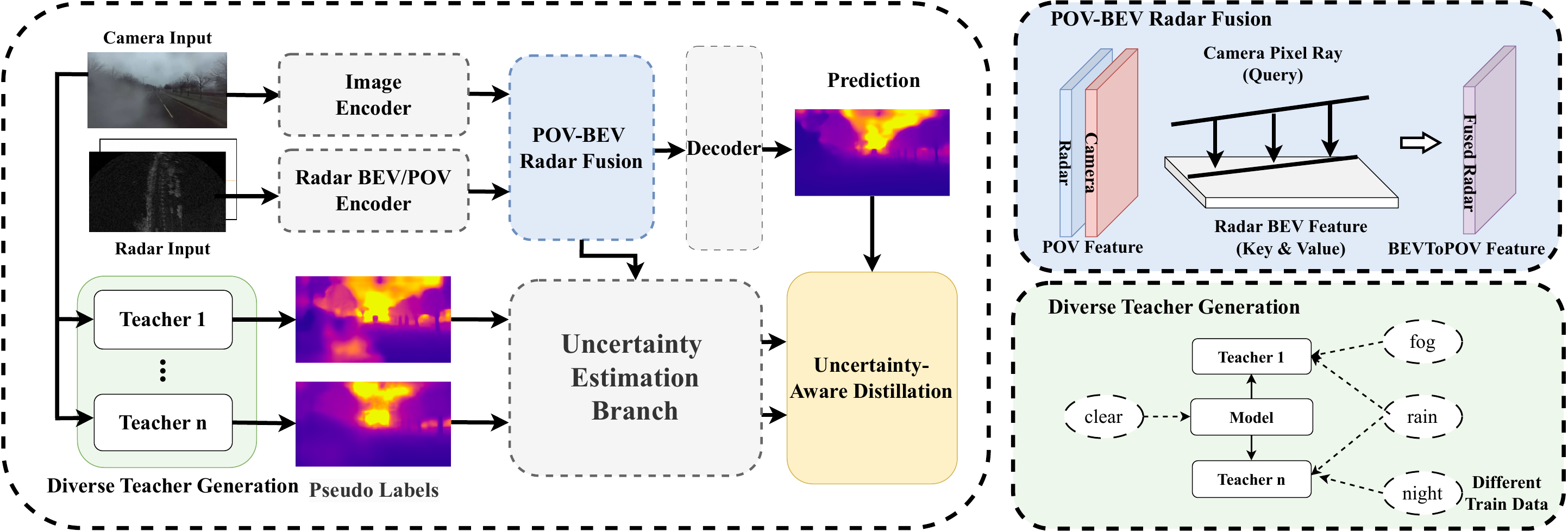}}
  \caption{An overview of our proposed method. Our approach first generates diverse teacher models, each of which is tailored to specific weather while introducing a novel distillation loss to improve knowledge transfer. Then the student model can estimate the confidence of the pseudo-labels generated by the teacher models, which is subsequently used in the Uncertainty-Aware Distillation. Additionally, we propose a camera-radar fusion method based on POV-BEV Cross-Attention to enhance robustness.}
  \label{fig:overview}
\end{figure*}

\noindent\textbf{Problem Formulation.}
For a given all-weather dataset \( X_c \) that includes both sequential camera data \(\{I_s \mid s \in \{t-1, t+1\}\}\) and radar data \(\{R_s \mid s \in \{t-1, t+1\}\}\), where the weather domain \( c \in C = \{ \text{clear}, \text{night}, \text{rain}, \text{fog}, \text{snow} \} \), the model needs to simultaneously estimate the depth of each frame \(\hat{D}_t\) and ego-pose to the next timestamp \(\hat{T}_{t \rightarrow t'}\). The model is optimized by an image reconstruction loss, which can be derived from the above estimations.

Fig.~\ref{fig:overview} presents an overview of our proposed method. Our approach consists of a training pipeline, Uncertainty-Aware Multi-Teacher Distillation (UAMTD), and a camera-radar fusion technique, POV-BEV Camera-Radar Fusion (PBCRF). 
Our method employs a novel distillation approach to address the suboptimal results and poor robustness issues posed by training on real all-weather data. Unlike prior work that leans heavily on synthetic adverse-weather pairs and therefore inherits a synthetic-to-real domain gap~\cite{gasperini2023robust}, we directly exploit real all-weather captures and harness the complementary cues of multiple specialized teachers and radar geometry. We first introduce a simple yet effective Single-Teacher Distillation (STD) method. Building upon this, we introduce the Uncertainty-Aware Multi-Teacher Distillation (UAMTD) method, which consists of two key components: 1) Diverse Teachers Generation and 2) Uncertainty-Aware Distillation. These components enable the generation of high-performance, diverse teacher models, while the student model selectively learns from the teachers based on uncertainty estimation, achieving the most robust performance. Our Diverse Teachers Generation strategy first generates a set of teacher models \( \Phi = \{ \Phi_1, \dots, \Phi_n \} \). These teacher models perform inference on \( X_c \) to produce pseudo-labels \( \bar{D} = \{ \bar{D}_1, \dots, \bar{D}_n \} \). Our Uncertainty Estimation Branch then generates uncertainty maps \( U = \{ U_1, \dots, U_n \} \), which serve as confidence estimations for the pseudo-label outputs from each teacher, where both \( U \) and \( \bar{D} \) are in \( \mathbb{R}^{H \times W} \). Finally, the student model learns from the teacher models through Uncertainty-Aware Distillation, incorporating \( U \) and \( \bar{D} \). Meanwhile, the camera features are fused with radar features from both POV and BEV perspectives for robustness enhancement through the proposed POV-BEV Camera-Radar Fusion (PBCRF) method. This method facilitates cross-perspective fusion of these modalities using camera-pixel ray constraints.

\subsection{Single-Teacher Distillation}\label{sec:self-distillation}~Given the image at timestamp \( t \) and the ego-pose transformation from \( t \) to \( t' \), \( \hat{T}_{t \rightarrow t'} \), we can synthesize the image at \( t' \) using the estimated depth \( \hat{D}_t \) as the following:
\begin{equation}
    I_{t' \rightarrow t}(\hat{D}_t) = I_{t'}\big\langle \text{proj}(\hat{D}_t, \hat{T}_{t \rightarrow t'}, K) \big\rangle,
\end{equation}
where \( K \) represents the camera intrinsics, and \(\text{proj}(\cdot)\) denotes the projection of the estimated 3D points back to the camera view. The self-supervised depth estimation is optimized using the following photometric reprojection loss~\cite{godard2019monodepth2}, which consists of a similarity loss (SSIM) and an \( L_1 \) loss:
\begin{align}
    \label{eq:ph}
    \mathcal{L}_{p} &= \alpha \cdot \frac{1 - \text{SSIM}(I_t, I_{s \rightarrow t}(\hat{D}_t))}{2} \notag \\
    &\quad + (1 - \alpha) \cdot \| I_t - I_{s \rightarrow t}(\hat{D}_t) \|_1,
\end{align}
where \( \alpha \) is typically set to 0.85. \(\mathcal{L}_{p}\) is derived under the brightness-consistency assumption of clear training conditions. Consequently, training directly on all-weather data \(X_c\), which does not adhere to this assumption, leads to significant errors. In adverse weather, scattering, absorption, and dynamic precipitations break the static-scene and constant-radiance premises of reprojection, so the photometric term \(\mathcal{L}_p\) yields misleading gradients precisely where depth cues are most needed. In contrast, a model trained on clear data \(X_{\text{clear}}\) performs better and demonstrates a certain degree of generalizability~\cite{gasperini2023robust}. Therefore, we can use this as a teacher model \(\Phi_i\) to generate pseudo-labels. These pseudo-labels provide a supervised signal and help prevent abnormal predictions in regions that violate the brightness-consistency assumption. The student model $T$ can then be optimized to learn from the teacher by minimizing the following similarity loss:
\begin{equation}
\label{eq:sim}
\mathcal{L}_{\text{sim}} = \frac{1}{N} \sum_{j=1}^N \frac{|\Phi_i(X_c)_j - T(X_c)_j|}{|\Phi_i(X_c)_j|},
\end{equation}
where \( N \) is the number of samples in a batch.

However, solely optimizing this similarity loss would result in a trivial student model that merely imitates \(\Phi_i\), which has not seen adverse weather data during training. Therefore, we allow the student model $T$ to learn from \( X_c \) through a self-supervised loss \( \mathcal{L}_p \). The two terms play complementary roles: \(\mathcal{L}_{\text{sim}}\) supplies a weather-robust anchor from the clear-domain teacher, while \(\mathcal{L}_{p}\) lets the student adapt to the actual appearance of \(X_c\) and recover details the teacher misses; balancing them avoids both over-reliance on a blind teacher and drift caused by corrupted photometric signals. The overall single-teacher distillation loss is then formulated as follows:
\begin{equation}
\label{eq:self-distillation}
\mathcal{L}_{\text{d}} = \mathcal{L}_{p} + \mathcal{L}_{sim}.
\end{equation}

\subsection{Uncertainty-Aware Multi-Teacher Distillation}\label{sec:multi-teacher-distillation}

Inspired by~\cite{wen2024class}, the student model achieves good performance by learning from multiple diverse teachers capable of generating high-performance pseudo labels. This raises two key questions: 1) \textit{how to generate such teachers}, and 2) \textit{to what extent the student model should rely on these teacher models}. To address these two questions, we propose a diverse teacher generation strategy by using a subset of the all-weather data, and introduce the uncertainty-aware distillation loss, allowing the student model to learn from the teacher model based on uncertainty.

\noindent\textbf{Diverse Teachers Generation.}\label{sec:diverse_teacher_generation} Our main idea is to construct diverse “weather experts” as multiple teacher models by differentiating the training data. Given a base model \(\Phi_{\text{clear}}\) trained on clear data, we select subsets \( X_{\text{c}_i} (c_i \subset C) \) from \( X_c \) as the training data for each teacher model \(\Phi_{i}\). We apply the Single-Teacher Distillation method to train each teacher model to prevent abnormal outputs across various weather conditions. By training on different subsets of adverse-weather data, we can obtain a set of weather-expert models as teachers. Concretely, we partition the adverse-weather portion of \(X_c\) into \(n\) disjoint groups (\emph{e.g.}, by weather condition or by random split) and train each \(\Phi_i\) on its own group together with the shared clear data, so that every teacher specializes in a particular appearance distribution yet remains anchored to the clear-domain teacher through STD. This deliberate diversification, rather than ensembling near-identical models, is what endows the student with broad all-weather coverage.

\noindent\textbf{Uncertainty-Aware Distillation.}  Previous depth estimation methods utilize various types of uncertainty, such as epistemic and aleatoric uncertainty, but they are derived from a single teacher~\cite{poggi2020uncertainty}. In contrast, we adopt a multi-stage uncertainty mining and refinement Uncertainty Estimation Branch in the depth estimation task following~\cite{zhu2023unsupervised}, which estimates uncertainty by considering both the student's features \(\mathcal{F} = \{ \mathcal{F}^1, \dots, \mathcal{F}^k \}\) and the quality of the teacher's pseudo-labels \(\bar{D} = \left\{ \bar{D}_1, \dots, \bar{D}_n \right\}\). Crucially, the reliability of a teacher's pseudo-label is not absolute but \emph{relative}: a teacher that is confident yet wrong on a given pixel should be down-weighted even if its individual estimate looks sharp. By feeding the concatenated outputs of \emph{all} \(n\) teachers into the UEB at each stage, the branch learns to compare teachers against one another and against the student's own features, so that disagreement among teachers---a strong indicator of confusing or corrupted regions---is reflected as high uncertainty. This joint, comparative assessment is what distinguishes our formulation from prior single-teacher uncertainty weighting~\cite{poggi2020uncertainty,kendall2017uncertainties}. 
The uncertainty map is estimated in a coarse-to-fine manner across multiple stages. At each stage \( t \), the coarse uncertainty map \( U_{i}^{t} \) for the \( i \)-th teacher model output is generated by concatenating the uncertainty maps from all teachers with the student features:
\begin{equation}
    U_{i}^{t+1} = T_{\text{ueb}}^t \left( \mathrm{Concat}\left( U_{1}^{t}, \cdots, U_{n}^{t}, \mathcal{F}^{t} \right) \right) + U_{i}^{t},
\end{equation}
where \( T_{\text{ueb}}^t(\cdot) \) incorporates non-local self-attention and the residual block with merge/split operation to aggregate the features~\cite{zhu2023unsupervised, wang2022multi}. For the initial stage \( t=0 \), \( U_{i}^{0} \) is initialized to \(\bar{D}\), as follows:
\begin{equation}
    U_{i}^{0} = T_{\text{ueb}}^0 \left( \mathrm{Concat}\left( \bar{D}_{1}, \cdots, \bar{D}_{n}, \mathcal{F}^{0} \right) \right) + \bar{D}_{i}.
\end{equation}
In supervised learning for depth estimation, the model typically estimates both depth \(\hat{D}\) and uncertainty\(\ U\) by optimizing the negative log-likelihood~\cite{kendall2017uncertainties} as follows:
\begin{equation}
    -\log p(D_{\text{gt}} | \hat{D}, U) = \frac{(D_{\text{gt}} - \hat{D})^2}{U^2} + \log U^2.
\end{equation}
By replacing \( D_{\text{gt}} \) with pseudo-labels and incorporating Equation~\ref{eq:sim}, we derive the final uncertainty-aware multi-teacher distillation loss as follows:
\begin{equation}
\mathcal{L}_{ud} = \mathcal{L}_{p} + \sum_{i=1}^{n} \left( \frac{\mathcal{L}_{\text{sim}_i}}{U_i} + \log U_i^2 \right),
\end{equation}
where \( n \) denotes the number of teacher models (set to match the number of weather experts generated in Sec.~\ref{sec:diverse_teacher_generation}). The term \( \log U_i^2 \) serves as a regularization component, ensuring that the uncertainty estimation remains within a reasonable range. In practice, we instead predict the log variance, \( U_{\log_i} = \log(U_i)^2 \), to enhance numerical stability~\cite{kendall2018multi}. Consequently, the above equation is reformulated as:
\begin{equation}
\mathcal{L}_{ud} = \mathcal{L}_{p} + \sum_{i=1}^{n} \left( \mathcal{L}_{\text{sim}_i} \exp(-U_{\log_i}) + \frac{1}{2} U_{\log_i} \right).
\label{eq:uncertainty-aware-multi-distill}
\end{equation}
Here, we extend uncertainty estimation to the \emph{joint} assessment of multiple teachers' pseudo-label quality. By accounting for inter-teacher interactions, Eq.~\ref{eq:uncertainty-aware-multi-distill} enables a selective distillation process. When uncertainty is high, the student model perceives the teacher model's predictions in this region as unreliable, allowing for larger deviations from the teacher and prioritizing the optimization of $\mathcal{L}_p$. Furthermore, the pixel-wise uncertainty estimation dynamically adjusts the similarity loss, particularly in regions where self-supervised loss assumptions are violated, while the photometric loss remains effective in regions where these assumptions hold.

\noindent\textbf{Uncertainty Estimation Branch training.}
The UEB has no explicit label and is learned end-to-end through the attenuated negative-log-likelihood loss in Eq.~\ref{eq:uncertainty-aware-multi-distill}. Minimizing this loss drives each $U_i$ up where a teacher's residual $\mathcal{L}_{\text{sim}_i}$ is large, so that $\exp(-U_{\log_i})$ down-weights unreliable pseudo-labels (selective distillation), while the $\tfrac{1}{2}U_{\log_i}$ term penalizes over-large uncertainty and prevents the head from collapsing. Because the UEB shares the student backbone's multi-stage features \(\mathcal{F}^t\), it receives gradients that simultaneously encourage accurate depth and honest uncertainty: when the student is forced to agree with a poor teacher, the residual grows and the corresponding \(U_i\) is pushed up, releasing the student to follow \(\mathcal{L}_p\) instead. In this way the branch is regularized implicitly by the consistency between teachers and self-supervision, requiring no extra ground truth or hand-crafted confidence measure.

\subsection{POV-BEV Radar Fusion}~\label{sec:RadarFusion}As shown in Figure~\ref{fig:intro}(b), radar data can be projected into either Point-of-View (POV) or Bird’s Eye View (BEV). We propose a POV-BEV Radar fusion method that leverages both perspectives, particularly the denser points in BEV, whereas prior work only performed fusion in POV. Quantitatively, averaged over 60 RADIATE scenes spanning all weather conditions (clear, night, rain, fog, snow), BEV radar provides \textbf{239$\times$} more non-zero points per frame than POV radar (98{,}840 vs.\ 413), supplying substantially denser geometric cues. Our PBCA exploits this density through ray-constrained attention that adaptively weights reliable radar returns and suppresses noise, and is further complemented by the sparse-but-accurate POV radar. Importantly, our POV-BEV Camera-Radar Fusion (PBCRF) adopts an \emph{opposite} cross-view direction from conventional Lift-Splat-Shoot (LSS) style fusion~\cite{philion2020lift}: instead of lifting image features into BEV for supervised detection, our method uses pixel rays to pull the denser BEV radar back into the POV space (camera as query, BEV as key/value). Because the self-supervised depth target lives in POV, fusing the dense BEV radar back into POV keeps the fused features consistent with the per-pixel reprojection loss, whereas the opposite LSS direction would misalign it with our POV objective. This fused representation is then further complemented by the sparse-but-accurate POV radar. We use lightweight ResNet-18 networks to encode the image \( I_s \) and the radar data \( R_s \) from both POV and BEV, obtaining the camera features \(\mathcal{F}_C \in \mathbb{R}^{C \times H_{\text{POV}} \times W_{\text{POV}}}\), the radar features in POV \( F_R^{\text{POV}} \in \mathbb{R}^{C \times H_{\text{POV}} \times W_{\text{POV}}}\), and the radar features in BEV \( F_R^{\text{BEV}} \in \mathbb{R}^{C \times H_{\text{BEV}} \times W_{\text{BEV}}}\). First, we fuse the camera features and radar BEV features using a POV-BEV Cross-Attention mechanism. Subsequently, we further fuse the features in the POV perspective.

Recent work shows that image features from POV space can be transformed to BEV space using the non-learning transformation method from \cite{philion2020lift}. This lift is fully differentiable and introduces no learnable parameters, making it inexpensive to invert for our backward (BEV$\to$POV) attention. Each pixel in the image can be regarded as a ray in 3D space by assuming each pixel corresponds to a depth vector of predefined dimension \(\{\delta, 2\delta, \dots, |D|\delta\}\), where \(|D|\) is the maximum range and \(\delta\) is the depth interval. This ray is then projected into the BEV space, enabling every location in the camera feature \(\varphi_{\text{Ci}} \in \mathbb{R}^{1 \times d}\) to be mapped to the nearest radar features \({\varphi}_{\text{Ri}} \in \mathbb{R}^{l_{\text{ray}} \times d}\), where \(l_{\text{ray}}\) denotes the number of radar features corresponding to the ray associated with each pixel. Next, we use the camera feature \(\varphi_{\text{Ci}}\) as the query, and the corresponding radar features \({\varphi}_{\text{Ri}}\) serve as the key and value in the Cross-Attention mechanism. The newly aggregated radar features are as follows:
\begin{equation}
\hat{\varphi}_{\text{R}_{i}} = \text{softmax} \left( \frac{\varphi_{\text{Ci}} {\varphi}_{\text{Ri}}^T}{\sqrt{d}} \right) {\varphi}_{\text{Ri}}.
\end{equation}
We gather all the aggregated features \(\hat{\varphi}_{\text{R}_{i}}\) to obtain \(\mathcal{F}_R^{\text{BEVtoPOV}} \in \mathbb{R}^{C \times H_{\text{POV}} \times W_{\text{POV}}}\), and then progressively fuse the features:
\begin{equation}
\mathcal{F}_\text{fusion} = f_\text{fusion}\left(\mathcal{F}_R^{\text{POV}}, f_\text{fusion}\left(\mathcal{F}_C, \mathcal{F}_R^{\text{BEVtoPOV}}\right)\right),
\end{equation}
\noindent where
\begin{equation}
f_\text{fusion} = \text{BN}\left(\text{Conv}\left(\text{Concat}(\cdot, \cdot)\right)\right).
\end{equation}

\subsection{Training and Inference}
We employ the single-teacher distillation loss, \( \mathcal{L}_d \) (Eq.~\ref{eq:self-distillation}), to train the teacher models. The student model is optimized using the uncertainty-aware multi-teacher distillation loss, \( \mathcal{L}_{ud} \) (Eq.~\ref{eq:uncertainty-aware-multi-distill}). During both the training and testing phases, we use camera and radar data as input and do not employ any additional supervision data. The teacher models \(\{\Phi_i\}\) are kept frozen after the Diverse Teachers Generation stage; only the student backbone and the Uncertainty Estimation Branch are updated end-to-end by \(\mathcal{L}_{ud}\), so the extra parameters introduced by UAMTD incur no cost at inference. During test time, the teachers are discarded, and the student takes camera-radar input alone, preserving the single-network, real-time deployment of standard self-supervised depth estimators.

\section{Experiment}
\subsection{Experimental Setup}

\textbf{Dataset.}~To evaluate algorithmic performance, our study utilizes two widely recognized multi-weather autonomous driving benchmarks: RADIATE~\cite{sheeny2021radiate} and nuScenes~\cite{caesar2020nuscenes}. Specifically, RADIATE provides 180 minutes of real-world vehicular logs across diverse environments such as city streets and highways. Although delivering precise 360-degree top-down radar representations, this dataset lacks Doppler velocity metrics and formal evaluation sets. Following the original creators' recommended partition, we discard stationary frames during training. Retaining motionless sequences severely compromises self-supervised depth predictions~\cite{gasperini2023robust}. Consequently, the dataset yields 63,359 training samples and 10,679 validation frames across \textit{sun(clear)} (4,957), \textit{night} (2,863), \textit{fog} (1,427), \textit{rain} (746), and \textit{snow} (686) conditions. Notably, snowy instances are completely absent from this training corpus. Conversely, nuScenes encompasses fifteen hours of navigating through Singapore and Boston, featuring complex environments under challenging weather. We categorize these situations into \textit{day-clear} (optimal visibility), \textit{nighttime} (incorporating rainy nights), and \textit{day-rain} conditions. Adopting the standard split from~\cite{gasperini2023robust}, our setup employs 15,129 synchronized sensor recordings for training. The 6,019-image verification portion contains 4,449 \textit{day-clear}, 602 \textit{night}, and 1,088 \textit{rain} examples.\\
\noindent\textbf{Compared Methods.}~To the best of our knowledge, training self-supervised depth estimation models directly on real-world all-weather data remains underexplored. Therefore, we establish a comprehensive benchmark comprising three main categories. \textbf{1) Standard Architecture Baselines.} To evaluate general performance and demonstrate the architecture-agnostic nature of our UAMTD method, we compare against widely adopted single-frame baselines (MonoDepth2~\cite{godard2019monodepth2}, Lite-Mono~\cite{zhang2023lite}, and MonoViT~\cite{zhao2022monovit}) as well as a recent temporal multi-frame network (ManyDepth2~\cite{zhou2025manydepth2}). \textbf{2) Robustness and Synthetic-Data Baselines.} To assess robustness under adverse conditions, we compare with RNW~\cite{wang2021regularizing}, which specifically leverages clear and night images. Furthermore, we expand our comparison to include recent state-of-the-art methods relying on synthetic adverse-weather data (e.g., md4all-DD~\cite{gasperini2023robust}, WeatherDepth~\cite{wang2024weatherdepth}, Syn2Real~\cite{yan2025synthetic}, and SEC-Depth~\cite{cao2026learning}) to evaluate performance and demonstrate the complementary benefits of our framework. \textbf{3) Camera-Radar Fusion Baselines.} To validate our camera-radar fusion module, we compare against two leading camera-radar fusion architectures evaluated from the Point of View (POV) perspective: R4Dyn~\cite{gasperini2021r4dyn} and CaFNet~\cite{sun2024cafnet}. Since their original training pipelines require additional supervisory signals (\emph{e.g.}, LiDAR or object boxes) unavailable in our purely self-supervised setting, we isolate and train their core radar fusion architectures using our framework to ensure a fair comparison.

\subsection{Evaluation Metrics}
\textbf{Metrics.}
We evaluate the performance of all models using a comprehensive set of metrics following prior work~\cite{wen2024class}. Specifically, we adopt two widely used error metrics, AbsRel and RMSE, along with the threshold-based precision metric $\delta < 1.25$. For all experiments, the maximum evaluation depth is capped at 80m to ensure a fair and consistent comparison across different methods and conditions. Unless otherwise specified, all experiments follow the standard self-supervised setting and report \emph{median-scaled} errors (scale aligned per frame via the median of the predicted depth).

\subsection{Implement Details}
We use AdamW as the optimizer, with hyperparameters set to ($\beta_{1}=0.9, \beta_{2}=0.999$) and set the initial learning rate to $1 \times 10^{-4}$, and the weight decay to $1 \times 10^{-2}$. All experiments are implemented using PyTorch, with a batch size of 12 and trained on a single NVIDIA 3090. When using the distillation technique, we freeze the PoseNet, following~\cite{woo2025prodepth}. Additionally, for all experiments, we apply brightness, saturation, contrast, hue jitter adjustments, and random horizontal flips the same as in \cite{zhang2023lite} as the data augmentation. For the RADIATE dataset, we use \{\( {\text{clear}}, {\text{rain}},  {\text{fog}} \)\} and \{\( {\text{clear}}, {\text{rain}}, {\text{night}} \)\} as the differentiated teacher (weather experts) training data with a resolution of $640 \times 320$. Specifically, due to the differing frame rates (FPS) of the two sensors, the number of camera-radar pairs is only about 1/4 of the number of camera images. Using only the camera-radar pairs to train the network is insufficient for convergence. Therefore, we alternate between using camera images and camera-radar pairs at each epoch. For the nuScenes dataset, it takes about 15 hours to achieve convergence results trained by our method. For the nuScenes dataset, we use \{\( {\text{clear}}, {\text{night}} \)\} and \{\( {\text{clear}},  {\text{rain}} \)\} as the differentiated teacher (weather experts) training data with an image size of $576 \times 320$, and it takes about 13.5 hours to achieve convergence results by preprocessing the images as binary files.




\subsection{Quantitative Comparisons}

\begin{table*}[!t]
\small
\caption{Comparison with existing state-of-the-art self-supervised depth estimation methods. \textit{tr.data} refers to the training data: `clear' indicates clear data of day, and `all' refers to all-weather data. \colorbox{gray!30}{Gray} indicates unacceptably high errors. Row 1 $\sim$ 12 compare our UAMTD method applied to different architectures (Monodepth2, MonoVit, ManyDepth2, and Lite-Mono). Row 13 $\sim$ 15 compare our camera-radar fusion method built upon UAMTD, where `ours$*$' denotes the application of the PBCRF method. The best results are highlighted in \textbf{bold}.}
\vspace{-3mm}
\setlength{\tabcolsep}{1.6pt}
\begin{center}
\resizebox{\textwidth}{!}{%
\begin{tabular}{ll|lll|lll|lll|lll|lll}
\toprule
& &  \multicolumn{3}{c|}{\textit{clear}} & \multicolumn{3}{c|}{\textit{night}} & \multicolumn{3}{c|}{\textit{rain}} & \multicolumn{3}{c|}{\textit{fog}} &
\multicolumn{3}{c}{\textit{snow}} \\
Method & tr.data & absRel↓ & RMSE↓ & $\delta_1$↑ & absRel↓ & RMSE↓ & $\delta_1$↑ & absRel↓ & RMSE↓ & $\delta_1$↑ & absRel↓ & RMSE↓ & $\delta_1$↑ & absRel↓ & RMSE↓ & $\delta_1$↑ \\
\midrule
Monodepth2~\cite{godard2019monodepth2} & \textit{clear}  & 0.2120 & 6.643 & 70.35 & 0.1822 & 6.052 & 74.14 & 0.2702 & 6.383 & 61.28 & 0.3265 & 6.741 & 57.46 & 0.6945 & 10.24 & 29.37 \\
Monodepth2\cite{godard2019monodepth2} & \textit{all} & 0.2258 & 6.762 & 66.78 & \cellcolor{gray!30}1.2211 & \cellcolor{gray!30}17.13 & \cellcolor{gray!30}14.13 & \textbf{0.2413} & 5.721 & 60.71 & 0.2965 & \textbf{3.662} & 45.15 & \textbf{0.5284} & \textbf{7.732} & \textbf{40.46} \\
Ours & \textit{all} & \textbf{0.2104} & \textbf{6.614} & \textbf{71.03} & \textbf{0.1647} & \textbf{5.788} & \textbf{77.37} & 0.2488 & \textbf{5.406} & \textbf{61.57} & \textbf{0.2832} & 4.544 & \textbf{58.12} & 0.6812 & 10.22 & 36.42 \\
\midrule

MonoVit~\cite{zhao2022monovit} & \textit{clear} & 0.1963 & 6.394 & 76.98 & 0.1626 & 5.744 & 76.94 & 0.2332 & 5.889 & 67.41 & 0.2884 & 4.586 & 54.80 & 0.5821 & 9.362 & 55.30  \\
MonoVit~\cite{zhao2022monovit} & \textit{all} & 0.2132 & 6.182 & 69.82 & \cellcolor{gray!30}0.9185 & \cellcolor{gray!30}12.62 & \cellcolor{gray!30}16.55 & 0.2274 & \textbf{5.073} & 67.05 & 0.2928 & \textbf{3.576} & 45.74 & \textbf{0.5430} & \textbf{7.572} & 33.17 \\
Ours & \textit{all} & \textbf{0.1843} & \textbf{5.935} & \textbf{77.58} & \textbf{0.1598} & \textbf{5.735} & \textbf{78.53} & \textbf{0.2116} & 5.162 & \textbf{69.71} & \textbf{0.2576} & 3.671 & \textbf{59.05} & 0.5577 & 9.124 & \textbf{59.27} \\
\midrule
ManyDepth2~\cite{zhou2025manydepth2} & \textit{clear} & 0.1780 & 5.797 & 77.70 & 0.1500 & \textbf{5.623} & 79.40 & 0.2560 & 5.712 & 62.90 & 0.2410 & 3.277 & 63.60 & 0.4600 & 7.333 & 55.90 \\
ManyDepth2~\cite{zhou2025manydepth2} & \textit{all} & 0.1890 & 5.798 & 77.30 & 0.2080 & 6.338 & 66.30 & 0.2590 & 6.121 & \textbf{65.10} & 0.3940 & 4.846 & 59.40 & 0.4780 & \textbf{7.068} & 47.60 \\
Ours & \textit{all} & \textbf{0.1690} & \textbf{5.417} & \textbf{78.30} & \textbf{0.1480} & 5.755 & \textbf{79.70} & \textbf{0.2330} & \textbf{5.331} & 65.00 & \textbf{0.2190} & \textbf{2.992} & \textbf{66.40} & \textbf{0.4530} & 7.513 & \textbf{57.00} \\
\midrule
Lite-Mono~\cite{zhang2023lite}  & \textit{clear} & 0.1802 & 6.271 & 79.27 & 0.1630 & 5.870 & 78.15 & 0.2616 & 6.475 & 64.45 & 0.3318 & 6.029 & 60.15 & 0.6426 & 11.14 & 52.46 \\
Lite-Mono~\cite{zhang2023lite} & \textit{all}  & 0.2133 & 6.450 & 71.67 & \cellcolor{gray!30}0.9498 & \cellcolor{gray!30}13.05 & \cellcolor{gray!30}15.71 & 0.2541 & 6.191 & 65.45 & 0.3802 & 4.356 & 39.14 & 0.7159 & 10.06 & 28.69 \\
Ours & \textit{all} & \textbf{0.1706} & \textbf{5.810} & \textbf{80.65} & \textbf{0.1462} & \textbf{5.628} & \textbf{81.06} & \textbf{0.2158} & \textbf{5.011} & \textbf{69.88} & \textbf{0.2514} & \textbf{3.706} & \textbf{63.31} & \textbf{0.5165} & \textbf{8.562} & \textbf{58.70} \\
\midrule
Ours + CaFNet~\cite{sun2024cafnet} & \textit{all} & 0.1749 & 5.916  & 79.86 & \textbf{0.1475} & \textbf{5.620} & \textbf{80.61} & 0.2194 & 5.102 & 69.09 & 0.2490 & 3.663 & 63.17  & 0.5597 & 9.165 & 57.23 \\
Ours + R4dyn~\cite{gasperini2021r4dyn} & \textit{all} & 0.1660 & 5.614 & 80.67 & 0.1478 & 5.722 & 80.10 & 0.2031 & 4.560 & 71.43 & 0.2414 & 3.453 & \textbf{63.39}  & 0.5115 & 8.371 & \textbf{58.37} \\
Ours$*$ & \textit{all}  & \textbf{0.1630} & \textbf{5.534} & \textbf{80.76} & 0.1482 & 5.805 & 79.96 & \textbf{0.1991} & \textbf{4.483} & \textbf{72.12} & \textbf{0.2400} & \textbf{3.391} & 63.19 & \textbf{0.5025} & \textbf{8.120} & 57.85 \\
\bottomrule
\end{tabular}
}
\end{center}
\label{table:radiata}
\end{table*}

\noindent\textbf{Camera-only Setting.} We evaluate our method by comparing it against two baselines: training on clear data only and directly training on all-weather data. As discussed before, directly training on all-weather data often degrades performance, especially at night, due to the violation of self-supervised loss assumptions. In contrast, our proposed UAMTD method effectively mitigates these issues, safely unlocking the potential of all-weather data across various architectures.

For single-frame architectures (Table~\ref{table:radiata}, rows 1 $\sim$ 12), UAMTD yields consistent improvements. For example, applying UAMTD to Lite-Mono reduces the absRel error by 20\%, 85\%, 15\%, 34\%, and 28\% across clear, night, rain, fog, and snow conditions compared to direct all-weather training, while also solidly outperforming the clear-only baseline by 5\%, 10\%, 18\%, 24\%, and 20\%, respectively. Although Monodepth2 and MonoVit with UAMTD show slightly higher errors in rare cases like snow compared to direct all-weather training, they successfully avoid the catastrophic failures that the direct approach suffers at night, demonstrating much stronger overall robustness.

Furthermore, we verify the architecture-agnostic nature of UAMTD by applying it to ManyDepth2~\cite{zhou2025manydepth2}, a recent multi-frame method that constructs cost volumes from consecutive frames. While ManyDepth2 attains competitive performance on clear data alone, direct all-weather training severely destabilizes it, with absRel rising from 0.150 to 0.208 at night and from 0.241 to 0.394 in fog. Applying UAMTD enables the student model to exploit all-weather data while anchoring to the robust clear teacher, reducing absRel by 5.1\%, 1.3\%, 8.9\%, 9.1\%, and 1.5\% across the five respective conditions over the clear-only baseline. This confirms that our distillation strategy seamlessly transfers to and enhances temporal multi-frame networks.

\noindent\textbf{Camera-Radar Fusion Setting}. We mainly compare our method with two state-of-the-art methods for camera-radar fusion in the Point of View perspective. Since the radar fusion method has only been studied with training on clear data and uses additional lidar or object boxes for supervision, which are not available in our experimental setting, we apply the UAMTD method and compare the efficacy of camera-radar fusion for a fair comparison. As shown in Table~\ref{table:radiata}, row 10 $\sim$ 12, our method achieves the best on the clear, rain, fog, and snow conditions, especially in the RMSE metric, by 6.5\%, 12.1\%, 7.4\% 11.4\%, and 1.4\%, 1.7\%, 1.8\% 3.0\% compared to CaFNet and R4dyn, respectively.

\begin{table*}[!t]
\small
\setlength{\tabcolsep}{3pt}
\caption{Comparisons of our method with existing state-of-the-art self-supervised depth estimation methods on the nuScenes dataset. The best performance is highlighted in \textbf{bold} and the second best is \underline{underlined}.}
\vspace{-3mm}
\begin{center}
\resizebox{0.8\textwidth}{!}{%
\begin{tabular}{ll|lll|lll|lll}
\toprule
&& \multicolumn{3}{c|}{\textit{day-clear}}
   & \multicolumn{3}{c|}{\textit{night}}
   & \multicolumn{3}{c}{\textit{day-rain}} \\
Method & tr.data
       & absRel↓ & RMSE↓ & $\delta_1$↑
       & absRel↓ & RMSE↓ & $\delta_1$↑
       & absRel↓ & RMSE↓ & $\delta_1$↑ \\
\midrule

Monodepth2~\cite{godard2019monodepth2}
& \textit{clear}
& \textbf{0.1374} & \textbf{6.694} & \underline{84.99}
& \underline{0.2829} & \underline{9.749} & \underline{51.89}
& \underline{0.1729} & \underline{7.749} & \underline{77.55} \\

Monodepth2~\cite{godard2019monodepth2}
& \textit{all}
& 0.1630 & 6.979 & \textbf{85.23}
& 2.4142 & 34.23 & 12.49
& 0.3302 & 9.152 & 67.96 \\

RNW~\cite{wang2021regularizing}
& \textit{clear \& night}
& 0.2872 & 9.185 & 56.21
& 0.3333 & 10.10 & 43.72
& 0.2952 & 9.341 & 57.21 \\

\textbf{Ours}
& \textit{all}
& \underline{0.1429} & \underline{6.785} & 84.87
& \textbf{0.2573} & \textbf{9.593} & \textbf{56.39}
& \textbf{0.1656} & \textbf{7.460} & \textbf{78.93} \\

\bottomrule
\end{tabular}%
}
\end{center}
\label{table:comp_nuscenes}
\end{table*}

\noindent\textbf{Results on nuScenes.} We further evaluate on the nuScenes dataset under the standard median-scaled self-supervised protocol (Table~\ref{table:comp_nuscenes}). Our method attains the best absRel at \textit{night} (0.2573) and \textit{day-rain} (0.1656), and remains competitive on \textit{day-clear} (0.1429, second only to the clear-trained Monodepth2 at 0.1374). The gains concentrate in adverse conditions: at night we reduce absRel by 9.1\% over the clear-trained Monodepth2 (0.2829) and by 22.8\% over RNW~\cite{wang2021regularizing} (0.3333), a night-specific method, while lifting $\delta_1$ from 51.89\% and 43.72\% to 56.39\%; under day-rain, absRel drops to 0.1656 with $\delta_1$ reaching 78.93\%. In contrast, training Monodepth2 directly on all-weather data collapses at night (absRel 2.4142). These results confirm that our method effectively exploits all-weather data and generalizes beyond the RADIATE benchmark.

\begin{table*}[!t]
\small
\caption{Comparison with recent robust self-supervised depth methods on nuScenes (day-clear / night / day-rain). All methods are evaluated in \textbf{metric scale} via velocity weak supervision (no median scaling), 80\,m cap---distinct from the median-scaled protocol used in Table~\ref{table:comp_nuscenes}. \textbf{Bold}: best; \underline{underline}: second best.}
\vspace{-3mm}
\begin{center}
\begin{tabular}{l|ccc|ccc|ccc}
\toprule
\multirow{2}{*}{Method} &
\multicolumn{3}{c|}{day-clear} & \multicolumn{3}{c|}{night} & \multicolumn{3}{c}{day-rain} \\
& aRel$\downarrow$ & RMSE$\downarrow$ & $\delta_1\uparrow$
& aRel$\downarrow$ & RMSE$\downarrow$ & $\delta_1\uparrow$
& aRel$\downarrow$ & RMSE$\downarrow$ & $\delta_1\uparrow$ \\
\midrule
md4all-DD~\cite{gasperini2023robust} & \underline{0.1366} & \textbf{6.45} & 84.60
& 0.1917 & 8.50 & 71.04 & \underline{0.1414} & \underline{7.23} & \underline{80.98} \\
WeatherDepth~\cite{wang2024weatherdepth} & 0.1384 & 6.71 & \textbf{84.90}
& 0.1942 & 8.54 & 70.30 & 0.1508 & 7.50 & 80.58 \\
Syn2Real~\cite{yan2025synthetic} & 0.1395 & \underline{6.54} & 84.31
& \underline{0.1846} & \textbf{8.17} & \underline{71.15} & 0.1429 & \textbf{7.17} & 80.84 \\
SEC-Depth~\cite{cao2026learning} & 0.1572 & 6.95 & 82.42
& 0.2008 & 8.86 & 67.95 & 0.1627 & 7.71 & 79.42 \\
\midrule
Ours (UAMTD, syn-teacher) & \textbf{0.1365} & 6.55 & \underline{84.63}
& \textbf{0.1835} & \underline{8.28} & \textbf{71.18}
& \textbf{0.1399} & 7.30 & \textbf{81.13} \\
\bottomrule
\end{tabular}
\end{center}
\vspace{-3mm}
\label{tab:comp_robust}
\end{table*}

\begin{figure}[!t]
  \centering
   \includegraphics[width=\linewidth]{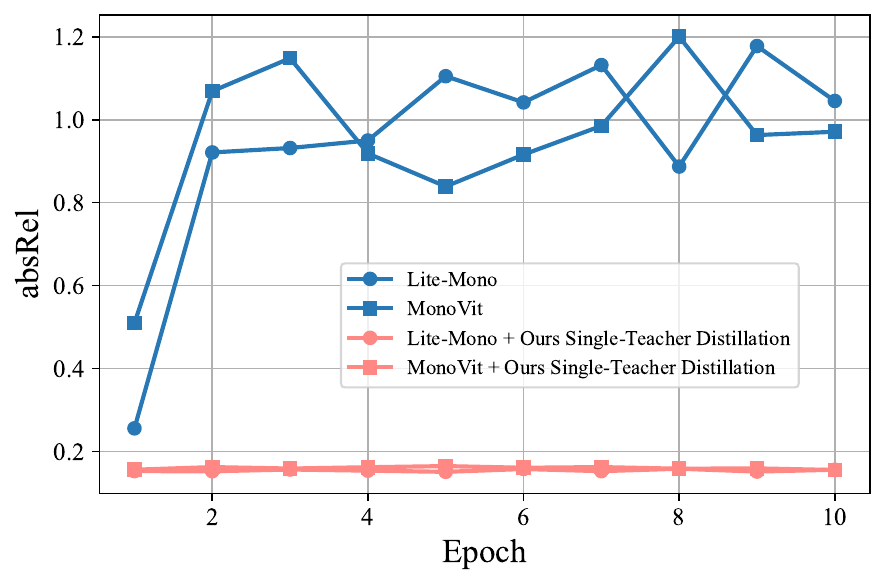}
   \vspace{-8mm}
    \caption{Trends in the change of absRel for different models on night conditions as the number of training epochs increases.}
    \label{fig:absRel_over_epochs}
    \vspace{-5mm}
\end{figure}

\noindent\textbf{Comparison with Synthetic-Data Training.} A prominent line of robust self-supervised depth estimation relies on synthetic adverse-weather data generated by image translation (\emph{e.g.}, GAN-based day$\rightarrow$night/rain), including md4all-DD~\cite{gasperini2023robust}, WeatherDepth~\cite{wang2024weatherdepth}, Syn2Real~\cite{yan2025synthetic}, and SEC-Depth~\cite{cao2026learning}. For a fair comparison against this family, Table~\ref{tab:comp_robust} evaluates all methods on nuScenes in \emph{metric scale} via velocity weak supervision (no median scaling) under the same MonoDepth2 backbone. Rather than treating synthetic data as an alternative to our framework, we show that the two are \emph{complementary}: we additionally train the weather-expert teachers of UAMTD on synthetic data following md4all-DD, while the student still learns from real all-weather data through uncertainty-aware distillation, leaving our core design unchanged. This additional synthetic supervision delivers the best absRel on all three conditions (\textit{day-clear} 0.1365, \textit{night} 0.1835, \textit{day-rain} 0.1399), surpassing purely synthetic-data methods and reducing the night absRel by 4.3\%, 5.5\%, and 8.6\% over md4all-DD, WeatherDepth, and SEC-Depth, respectively. This indicates that our distillation not only exploits real all-weather data but also seamlessly absorbs synthetic data to further boost robustness.

\noindent\textbf{Robustness and Generalization.} Our approach remains robust across unseen conditions and architectures. \emph{First}, although snow is excluded from training by design due to its scarcity~\cite{sheeny2021radiate}, our method still generalizes to it, reducing the snow absRel by 27.9\% over the all-weather-trained baseline and 19.6\% over the clear-trained baseline. \emph{Second}, our strategy is inherently architecture-agnostic, consistently improving both single-frame models (\emph{e.g.}, reducing Lite-Mono's night absRel by 85\% compared to direct all-weather training) and temporal multi-frame networks (\emph{e.g.}, improving ManyDepth2 by up to 9.1\% in fog over the clear-only baseline) without requiring any structural modifications.

\begin{table*}[!t]
\small
\caption{Comprehensive ablation studies of the proposed method. From top to bottom, the blocks evaluate: (1) the Uncertainty-Aware Multi-Teacher Distillation (Rows 1--5); (2) comparisons with alternative uncertainty estimation methods (Rows 6--7); (3) the effectiveness of radar fusion strategies (Rows 8--11); (4) the impact of teacher model selection based on different weather conditions (where c, r, f, and n denote clear, rain, fog, and night, respectively) (Rows 12--15); and (5) comparative analysis on modality balancing (Rows 16--17). Best and second-best results are highlighted in \textbf{bold} and \underline{underlined}, respectively.}
\vspace{-5mm}
\setlength{\tabcolsep}{1.6pt}
\begin{center}
\resizebox{\textwidth}{!}{%
\begin{tabular}{ l |ccc|ccc|ccc|ccc|ccc}
\toprule
& \multicolumn{3}{c|}{\textit{clear}} & \multicolumn{3}{c|}{\textit{night}} & \multicolumn{3}{c|}{\textit{rain}} & \multicolumn{3}{c|}{\textit{fog}} &
\multicolumn{3}{c}{\textit{snow}} \\
Method & absRel↓ & RMSE↓ & $\delta_1$↑ & absRel↓ & RMSE↓ & $\delta_1$↑ & absRel↓ & RMSE↓ & $\delta_1$↑ & absRel↓ & RMSE↓ & $\delta_1$↑ & absRel↓ & RMSE↓ & $\delta_1$↑\\
\midrule
STD  & \underline{0.1730} & 5.883 & 79.68 & 0.1511 & 5.707 & 80.14 & 0.2318 & 5.444 & 67.38 & 0.2566 & 3.786 & 62.23 & \textbf{0.4880} & \textbf{8.205} & \textbf{59.51}\\
Teacher: (c,r,n) ~& 0.1739 & \underline{5.873} & 79.70 & 0.1568 & 5.769 & 78.94 & 0.2290 & 5.407 & 67.84 & 0.2605 & 3.850 & 61.61 & 0.5581 & 9.176 & 56.97 \\
Teacher: (c,r,f) ~& 0.1769 & 6.053 & \underline{80.37}  & 0.1500 & 5.628 & 80.58 & \underline{0.2238} & 5.321 & \underline{69.15} & \textbf{0.2499} & 3.824 & \textbf{63.68} & 0.5226 & 8.611 & 56.43  \\
Average MTD ~  & 0.1770 & 6.000 & 79.35 & \underline{0.1482} & \textbf{5.619} & \textbf{81.17} & 0.2261 & \underline{5.160} & 68.27 & \underline{0.2510} & \textbf{3.675} & 62.77 & 0.5347 & \underline{8.526} & 53.34\\
UAMTD (ours)~ & \textbf{0.1706} & \textbf{5.810} & \textbf{80.65} & \textbf{0.1462} & \underline{5.628} & \underline{81.06} & \textbf{0.2158} & \textbf{5.011} & \textbf{69.88} & 0.2514 & \underline{3.706} & \underline{63.31} & \underline{0.5165} & 8.562 & \underline{58.70} \\ 
\midrule
MTD+Aleatoric Uncertainty~ & 0.1708 & \textbf{5.756} & 79.79 & 0.1663 & 6.047 & 77.38 & 0.2307 & 5.289 & 66.42 & 0.2541 & \textbf{3.420} & 61.87 & \textbf{0.4897} & \textbf{7.828} & 58.02 \\
UAMTD (ours)~ & \textbf{0.1706} & 5.810 & \textbf{80.65} & \textbf{0.1462} & \textbf{5.628} & \textbf{81.06} & \textbf{0.2158} & \textbf{5.011} & \textbf{69.88} & \textbf{0.2514} & 3.706 & \textbf{63.31} & 0.5165 & 8.562 & \textbf{58.70} \\
\midrule
w/o radar~ & 0.1706 & 5.810 & 80.65 & \textbf{0.1462} & \textbf{5.628} & \textbf{81.06} & 0.2158 & 5.011 & 69.88 & 0.2514 & 3.706 & 63.31 & 0.5165 & 8.562 & \textbf{58.70} \\
POV  & \underline{0.1660} & \underline{5.614} & \underline{80.67} & \underline{0.1478} & 5.722 & 80.10 & \underline{0.2031} & \underline{4.560} & \underline{71.43} & \underline{0.2414} & \underline{3.453} & \underline{63.39} & \underline{0.5115} & \underline{8.371} & \underline{58.37} \\
BEV & 0.1696 & 5.789 & 80.43 & 0.1486 & \underline{5.661} & \underline{80.11}  & 0.2187 & 5.061 & 69.23  & 0.2478 & 3.631 & \textbf{63.40} & 0.5259 & 8.683 & 56.80 \\
POV + BEV  & \textbf{0.1630} & \textbf{5.534} & \textbf{80.76} & 0.1482 & 5.805 & 79.96 & \textbf{0.1991} & \textbf{4.483} & \textbf{72.12} & \textbf{0.2400} & \textbf{3.391} & 63.19 & \textbf{0.5025} & \textbf{8.120} & 57.85 \\
\midrule
Teacher: (c,r,f)+(c,r,n)   
& \textbf{0.1706} & \underline{5.810} & \textbf{80.65} 
& \textbf{0.1462} & \underline{5.628} & \textbf{81.06} 
& \textbf{0.2158} & \textbf{5.011} & \textbf{69.88} 
& \underline{0.2514} & \underline{3.706} & \textbf{63.31} 
& \textbf{0.5165} & \textbf{8.562} & \textbf{58.70} \\
Teacher: (c,r,f)+(c,f,n) 
& 0.1790 & 6.038 & 79.27 
& 0.1534 & 5.677 & 79.29 
& 0.2270 & 5.214 & 67.68 
& \textbf{0.2495} & \textbf{3.538} & \underline{62.92} 
& 0.5607 & 9.036 & 52.29 \\
Teacher: (c,r,n)+(c,f,n) 
& 0.1732 & 5.932 & 80.20 
& 0.1487 & 5.665 & \underline{80.94} 
& 0.2256 & 5.277 & 68.31 
& 0.2535 & 3.747 & 62.66 
& \underline{0.5258} & \underline{8.584} & 56.48 \\
Teacher: (c,r)+(c,f)+(c,n)~ 
& \underline{0.1717} & \textbf{5.804} & \underline{80.23} 
& \underline{0.1478} & \textbf{5.627} & 80.43 
& \underline{0.2214} & \underline{5.147} & \underline{68.94} 
& 0.2556 & 3.794 & 62.44 
& 0.5496 & 9.078 & \underline{56.87} \\
\midrule
camera+radar (ours) & \textbf{0.1630} & \textbf{5.534} & 80.76 & \textbf{0.1482} & 5.805 & \textbf{79.96} & \textbf{0.1991} & \textbf{4.483} & \textbf{72.12} & \textbf{0.2400} & \textbf{3.391} & \textbf{63.19} & \textbf{0.5025} & \textbf{8.120} & 57.85 \\
camera+radar + AWC~\cite{qi2025towards} & 0.1650 & 5.659 & \textbf{80.80} & 0.1493 & \textbf{5.804} & 79.34 & 0.2083 & 4.675 & 70.43 & 0.2451 & 3.498 & 62.76 & 0.5238 & 8.548 & \textbf{58.37} \\
\bottomrule
\end{tabular}%
}
\end{center}
\label{table:ablation}
\end{table*}

\subsection{Ablation Study}
\label{sec:ablation}

\textbf{Distillation Learning.}~\textbf{1) Effect of Single-Teacher Distillation (STD)}  
As shown in Table~\ref{table:radiata} (row 1), training with STD on the all-weather dataset significantly improves performance across various weather conditions. As illustrated in Fig.~\ref{fig:absRel_over_epochs}, training directly on all-weather data makes the night absRel oscillate between 0.26 and 1.18, whereas STD stabilizes it at about 0.15 (more than 6$\times$ lower), confirming that anchoring the student to the robust clear teacher suppresses abnormal predictions where photometric consistency is violated.~\textbf{2) Effect of Uncertainty-Aware Multi-Teacher Distillation (UAMTD)}  
Compared to the direct STD method, our proposed UAMTD achieves a reduction in RMSE error by 1.2\%, 1.4\%, 8.0\%, and 2.1\% in clear, night, rain, and fog conditions, respectively. Thus, UAMTD offers a more robust and generalized learning strategy by dynamically leveraging teacher models based on their confidence levels. This ensures that the student model benefits from the most reliable information in different weather conditions, making UAMTD a superior choice overall.

\begin{figure}[!t]
\centering
\includegraphics[width=0.85\linewidth]{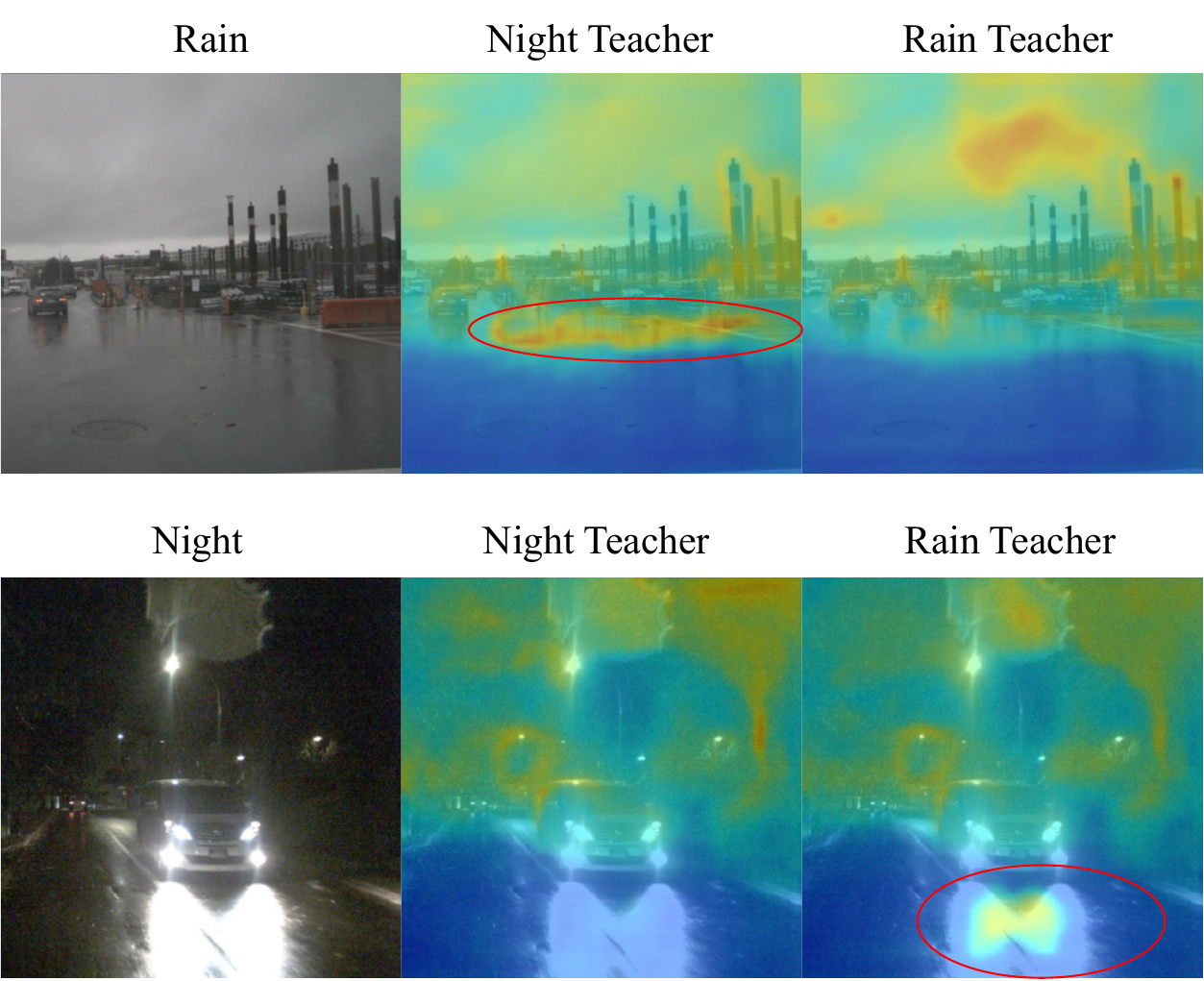}
\vspace{-4mm}
\caption{Comparison of uncertainty estimation across different teacher models in heatmap form. We marked the regions with significant uncertainty differences with red circles.}
\vspace{-5mm}
\label{fig:unc_comp}
\end{figure}

\noindent\textbf{Effect of Uncertainty Estimation in UAMTD} In Table~\ref{table:ablation}, we present an ablation study for the proposed UAMTD method. The table includes results for the individual teacher models (row~2 $\sim$ row~3) and Multi-Teacher Distillation (MTD), which simply averages the outputs of two teachers without considering uncertainty (Average MTD, row 4). The results show that the absRel and $\delta_1$ values in row 4 do not surpass those of the teacher models in clear, rain, and fog conditions. However, our proposed UAMTD method (row 5) outperforms the teacher models across all three metrics in clear, night, rain, and snow conditions and achieves a better RMSE than both teachers in fog. This demonstrates that estimating the confidence of teacher outputs and selectively learning from them provides a clear advantage. Furthermore, Fig.~\ref{fig:unc_comp} illustrates the interpretability of uncertainty estimation by comparing the uncertainty maps predicted by different teachers on the nuScenes dataset. The teacher trained on night data exhibits higher uncertainty on wet roads during rainy days, while the teacher trained on rain data shows higher uncertainty under varying nighttime lighting conditions. As a result, the student model learns to ignore high-uncertainty regions, leading to more effective knowledge distillation.


\noindent \textbf{Teacher Model Selection.} To ensure comprehensive robustness, the chosen teacher ensemble must cover all available adverse weather conditions in the training set: rain ($r$), fog ($f$), and night ($n$) (snow is excluded due to the lack of training data). Consequently, the ensemble requires a minimum of two and a maximum of three specialized teachers. As reported in Table \ref{table:ablation} (Rows 12--15), we experimented with several configurations, including the three-teacher setup $(c,r) + (c,f) + (c,n)$ and different two-teacher pairings. The results indicate that the $(c,r,f) + (c,r,n)$ combination yields the best overall performance. For instance, compared to the three-teacher setup, our configuration reduces the absRel error from 0.2214 to 0.2158 under rain conditions, and significantly improves the $\delta_1$ accuracy from 56.87\% to 58.70\% on the unseen snow test set. We hypothesize that this specific two-teacher configuration avoids the data fragmentation caused by dividing the dataset into three smaller subsets, thereby providing sufficient diversity and sample size for each teacher to learn robust representations. Therefore, we utilize $(c,r,f) + (c,r,n)$ as our optimal choice.

\noindent\textbf{Comparison with Other Uncertainty Estimation Methods.} Another uncertainty-weighting method used in distillation is aleatoric uncertainty~\cite{poggi2020uncertainty}. This approach views the model output as a Gaussian distribution, where the output variance represents the aleatoric uncertainty, capturing the model's uncertainty regarding data noise. Unlike our method, this does not require information exchange between the teacher and student models. As shown in Table~\ref{table:ablation} (row 6 $\sim$ 7), the results of the aleatoric uncertainty estimation method are worse than our method across all conditions except snow. However, its predictions are rather vague. More comparisons refer to our supplementary.
\begin{figure*}[!t]
  \centering
   \includegraphics[width=\linewidth]{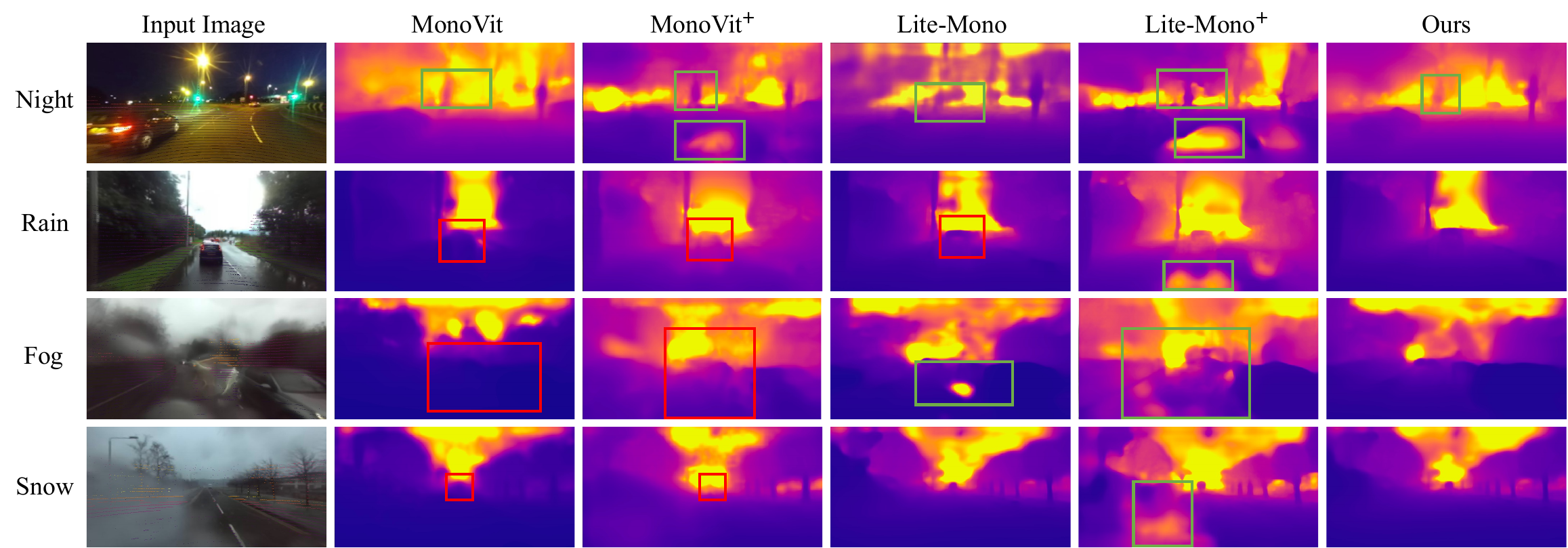}
   \vspace{-4mm}
    \caption{Visualization comparison with different methods. Green boxes highlight erroneous depth estimates, such as predicting the ground as a hole in rainy conditions. Red boxes indicate ambiguous object estimates, such as predicting the car and background with the same depth. Lite-Mono and MonoVit refer to models trained on clear data, while $\text{Lite-Mono}^{+}$ and $\text{MonoVit}^{+}$ refer to models trained on all-weather data. Our model is built on Lite-Mono and trained on all-weather data.
}
    \label{fig:visual_samples}
    \vspace{-3mm}
\end{figure*}

\noindent\textbf{Camera-Radar Fusion Analysis.}~In Table~\ref{table:ablation} (rows 8–11), simply fusing radar in POV or BEV results in only marginal performance improvement. We can observe that the POV radar alone is highly sparse: averaged over 60 RADIATE scenes spanning all weather conditions. BEV radar provides 239$\times$ more non-zero points per frame than POV (98{,}840 vs.\ 413); our PBCA is designed to exploit precisely this denser BEV signal via ray-constrained attention while complementing it with the sparse-but-accurate POV radar. In contrast, our method achieves the best performance across clear, rain, fog, and snow conditions, with RMSE reductions of 5\%, 11\%, 8\%, and 5\%, respectively. However, all radar fusion methods exhibit performance degradation at night compared to the camera-only approach, due to the quality of night scenes in the training set and the domain gap between the training and test sets.

\noindent\textbf{Efficiency Analysis.} As reported in Table~\ref{tab:efficiency}, the deployed camera-only sub-network runs at 8.78M parameters and 12.8\,ms (78\,FPS). The two teacher models and the uncertainty-mining branch are training-only and add zero inference cost, while the BEV-sampling cross-attention (PBCRF) adds 23.8M parameters and raises latency to 26.9\,ms (37\,FPS). Both configurations satisfy real-time onboard requirements.

\begin{table}[!t]
\small
\caption{Efficiency of the proposed components measured on a single NVIDIA V100 with a $1\times3\times320\times640$ input. The camera-only sub-network is shared by all same-backbone baselines and our inference. The teacher models and the uncertainty-mining branch are training-only and incur zero inference cost, as they are discarded after distillation.}
\label{tab:efficiency}
\centering
\setlength{\tabcolsep}{4pt}
\begin{tabular}{l|ccc}
\toprule
Method & \#Params (M) & Latency (ms) & FPS \\
\midrule
baseline (Monodepth2) & 8.78 & 12.8 & 78 \\
\;+\,Uncertainty branch & 8.78 (+0) & 12.8 (+0) & 78 (-0) \\
\;+\,PBCRF (radar fusion) & 32.6 (+23.8) & 26.9 (+14.1) & 37 (-41) \\
\bottomrule
\end{tabular}
\end{table}

\subsection{Qualitative Comparisons} 
Fig.~\ref{fig:visual_samples} illustrates the visual differences between our method and others. Our method demonstrates more precise depth predictions and fewer abnormal regions affected by adverse conditions. Additionally, it accurately predicts ground surfaces during nighttime. It effectively identifies vehicles and their edges on rainy and foggy days, as well as the trunks of trees and trains on the horizon on snowy days. Specifically, under foggy conditions (row 3), models like MonoVit and Lite-Mono, when trained on clear data, produce very blurred predictions with meaningless highlights. When trained on all-weather data, these models exhibit strange textures affected by raindrops. In contrast, our method accurately predicts the depth of vehicles with clear contours. Furthermore, we present a failure case in the first row, where our method incorrectly predicts the glare from a traffic light as a solid object. Addressing the physical properties of objects may help resolve this issue in future work.
\begin{figure}[!h]
  \centering
   \includegraphics[width=0.9\linewidth]{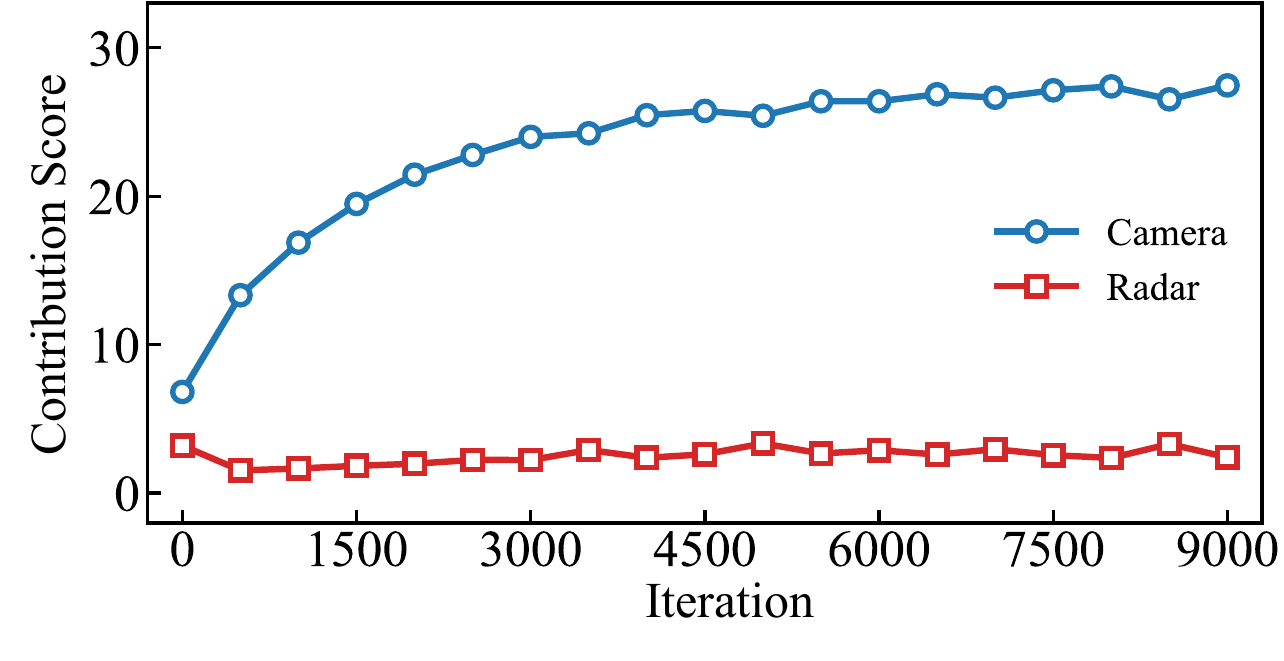}
   \vspace{-3mm}
    \caption{Visualization of the contribution scores for the camera and radar modalities.}
    \label{fig:modal_con}
    \vspace{-5mm}
\end{figure}

\section{Discussion}

\noindent\textbf{Analysis of Modality Imbalance.}
To accurately quantify the modality imbalance inherent in our camera-radar fusion framework, we adopt the Shapley-based contribution assessment~\cite{qi2025towards}, a game-theoretic approach that provides a rigorous metric for evaluating individual sensor importance during training. As explicitly visualized in the training dynamics of Figure~\ref{fig:modal_con}, the learning process exhibits a severe modality collapse. The network rapidly gravitates toward the dense visual cues of the camera branch, causing its contribution score to soar early in the optimization. Conversely, the radar's contribution plummets to near zero, indicating that the robust, weather-invariant physical properties captured by the radar are severely underutilized. 

To proactively alleviate this imbalance, we further applied the Adaptive Weight Constraint (AWC) regularization~\cite{qi2025towards}, which is designed to dynamically penalize the dominant camera branch and encourage the network to learn from the weaker radar stream. However, as shown in Table~\ref{table:ablation} (Rows 16--17), AWC yields highly inconsistent results across different environments. While it marginally improves performance in the Snow scenario ($\delta_1$: 58.37\% \vs 57.85\%), likely because radar offers critical compensation when visual textures are heavily corrupted, it degrades evaluation metrics in Night and Rain. We fundamentally attribute this failure to the drastic information capacity gap between the two modalities. Unlike experimental settings with more balanced multi-sensor configurations~\cite{qi2025towards}, over-constraining the camera here severely bottlenecks the acquisition of primary dense depth cues. Because the radar point cloud is extremely sparse and lacks high-resolution spatial context, it simply cannot independently compensate for the artificial suppression of the visual modality. Thus, simply transplanting existing balancing strategies is ineffective for camera-radar depth estimation. Designing specialized, capacity-aware architectures to effectively exploit their complementary strengths without arbitrarily penalizing the stronger modality remains a critical future direction.

\section{Conclusion}
In this paper, we presented an uncertainty-aware multi-teacher distillation method that significantly improved the depth estimation performance across various weather conditions when trained with all-weather data. Additionally, we showed a novel POV-BEV camera-radar fusion method. Extensive experiments on the widely-adopted real all-weather dataset demonstrated the effectiveness of our proposed method. In the future, we will apply our method into the real autonomous driving safety system.

\bibliographystyle{IEEEtran}
\bibliography{main}

\clearpage
\appendices
\section{Overview}
In the supplementary materials, we first provide a detailed explanation of the metric definitions in Section~\ref{sec:metric_def}. Section~\ref{sec:radar_fusion_comp} describes the implementation details of the compared camera-radar fusion method. In Section~\ref{sec:alea_comp}, we present a qualitative comparison between aleatoric uncertainty and our proposed uncertainty estimation method within the Uncertainty-Aware Distillation framework. Section~\ref{sec:md4all_comp} provides a comparison with the method using paired synthetic data. Finally, in Section~\ref{sec:further_qual_comp}, we present additional qualitative results demonstrating the superiority of the proposed method.

\section{Metric Definition}
\label{sec:metric_def}
\textbf{AbsRel} measures the error between the predicted values and the ground truth, normalized by the ground truth. Given the predicted depth \( \hat{D} \) and the ground truth \( D \), it is defined as:
\[
\text{AbsRel} = \frac{1}{N} \sum_{i=1}^N \frac{|\hat{D}_i - D_i|}{D_i},
\]
where \( N \) is the number of all pixels.\\
\noindent\textbf{RMSE} measures the root-mean-square error between the predicted depth values and the ground truth. It is defined as:
\[
\mathrm{RMSE} = \left(\frac{1}{N} \sum_{i=1}^N (\hat{D}_i - D_i)^2 \right)^{\frac{1}{2}},
\]
where \(N\) is the total number of pixels.\\
\noindent\textbf{Accuracy} measures the proportion of predictions whose error is within a specific threshold:
\[
\text{Accuracy}_t = \frac{1}{N} \sum_{i=1}^N \mathbf{1} \left( \delta_i < threshold \right),
\]
\[
\delta_i = \max \left( \frac{\hat{D}_i}{D_i}, \frac{D_i}{\hat{D}_i} \right),
\]
where \( \mathbf{1} \) is the indicator function.

\section{Comparison with Other Camera-Radar Fusion Methods}
\label{sec:radar_fusion_comp}
We compare our method with two state-of-the-art camera-radar fusion methods in the Point of View (POV) perspective. R4Dyn~\cite{gasperini2021r4dyn} employs the late fusion approach from~\cite{lin2020depth} to integrate radar inputs and utilizes 2D bounding boxes to introduce a novel radar supervision loss. However, since such supervision is unavailable in our setting, we compare only the radar fusion network with R4Dyn. CaFNet~\cite{sun2024cafnet} is a recent two-stage supervised camera-radar depth completion method. It leverages a multi-layer fusion strategy and a gated mechanism~\cite{singh2023depth} for radar fusion. However, its first-stage confidence map generation requires ground truth supervision, which is incompatible with a self-supervised setting. Therefore, we compare only the radar fusion network of CaFNet, excluding the confidence map.

\begin{table*}[!t]
\centering
\caption{Comparison with Paired Synthetic Data.}
\begin{tabular}{ll|cc|cc|cc}
\toprule
&& \multicolumn{2}{c|}{\textit{day-clear} -- nuScenes}
   & \multicolumn{2}{c|}{\textit{night} -- nuScenes}
   & \multicolumn{2}{c}{\textit{day-rain} -- nuScenes} \\
Method & tr.data
       & absRel & $\delta_1$
       & absRel & $\delta_1$
       & absRel & $\delta_1$ \\
\midrule

md4all
& \textit{clear+synthetic}
& \textbf{0.1393} & 84.08
& \textbf{0.2335} & \textbf{66.26}
& 0.1663 & 78.30 \\

\textbf{Ours UAMTD}
& \textit{all}
& 0.1429 & \textbf{84.87}
& 0.2573 & 56.39
& \textbf{0.1656} & \textbf{78.93} \\
\bottomrule
\end{tabular}%
\label{table:md4all_comp}
\end{table*}

\begin{figure*}[!tbp]
  \centering
   \includegraphics[width=0.6\linewidth]{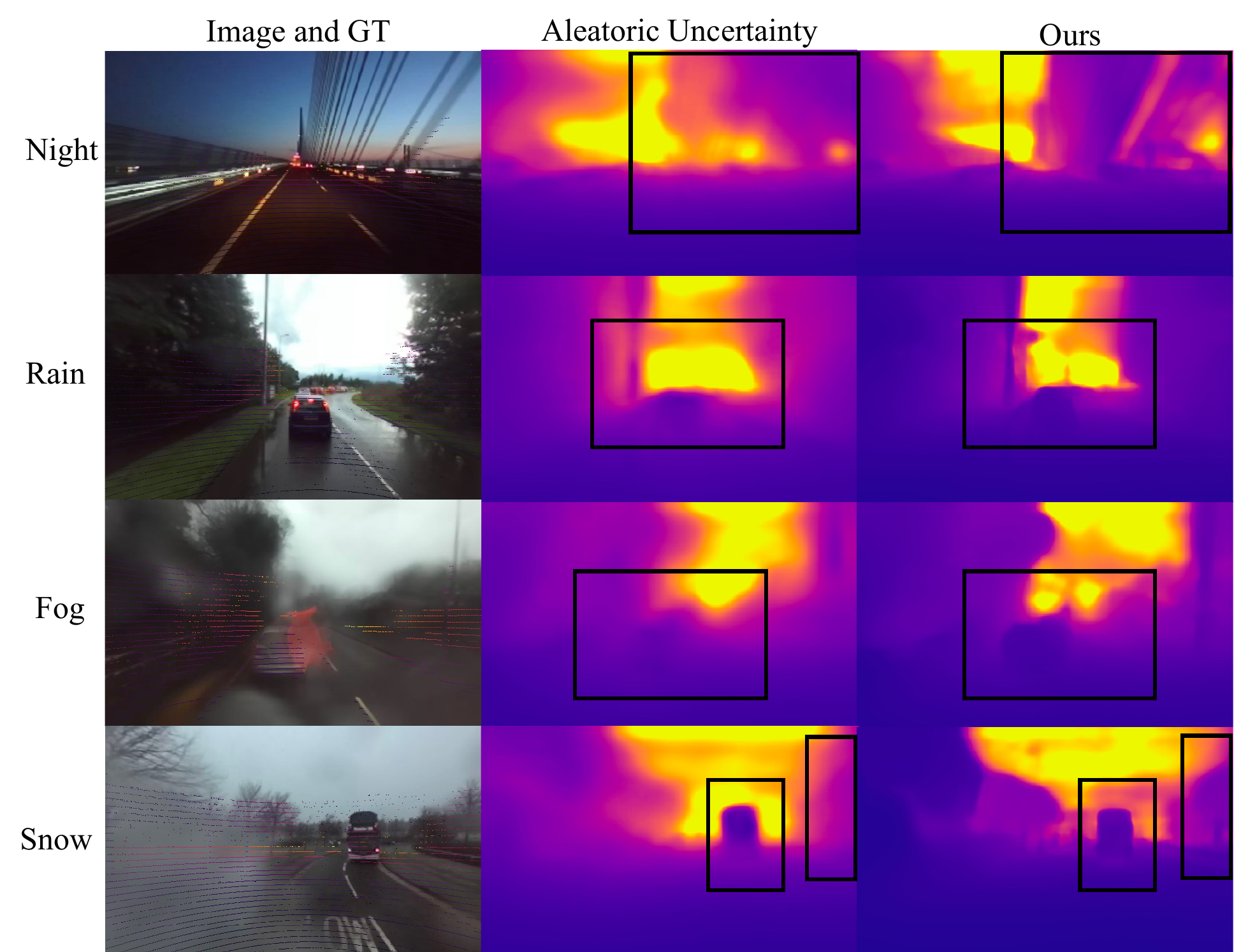}
    \caption{Visualization comparison of aleatoric uncertainty and our proposed uncertainty. Our method provides more detailed features, highlighted by the boxes.}
    \label{fig:uncer_comp}
\end{figure*}

\section{Comparison with Methods Based on Aleatoric Uncertainty}
\label{sec:alea_comp}
We compare the results of using our proposed uncertainty with the widely used aleatoric uncertainty~\cite{Guo2024UNIKD} in the Uncertainty-Aware Multi-Teacher Distillation. As shown in Fig.~\ref{fig:uncer_comp}, the results obtained using aleatoric uncertainty are relatively blurry, whereas our depth predictions contain more detailed features. In the Ablation Study section, aleatoric uncertainty outperforms our method under snow conditions but may have limited practical value due to poor visualization results.

\section{Comparison with Methods Based on Paired Synthetic Data}
\label{sec:md4all_comp}
Comparing our method with synthetic data-based approaches like md4all~\cite{gasperini2023robust} is inherently unfair, as synthetic datasets typically outnumber real ones. For instance, md4all includes 89K adverse-weather synthetic samples but only 14K real samples---over six times fewer. The results in Table~\ref{table:md4all_comp} compare methods using ResNet-18 as the backbone, relying solely on camera data to maintain a fully self-supervised setting. While our method underperforms md4all at night---an expected outcome given the disparity in data---it achieves the best performance in day-rain scenarios. This highlights the effectiveness of leveraging limited real-weather data, demonstrating greater adaptability through the combination of similarity loss and photometric loss. Notably, training our method on the nuScenes dataset, including diverse teacher training, requires only 13.5 hours on a single V100. In contrast, synthetic data-based methods like md4all, which involve training additional models (\emph{e.g.}, ForkGAN) on datasets such as BDD100K and nuScenes, can take over 25 hours (To optimize efficiency, we precompute resized images, store them as binary files, and apply code optimizations based on the public training framework, significantly reducing training time).

\begin{figure*}[htbp]
  \centering
   \includegraphics[width=0.6\linewidth]{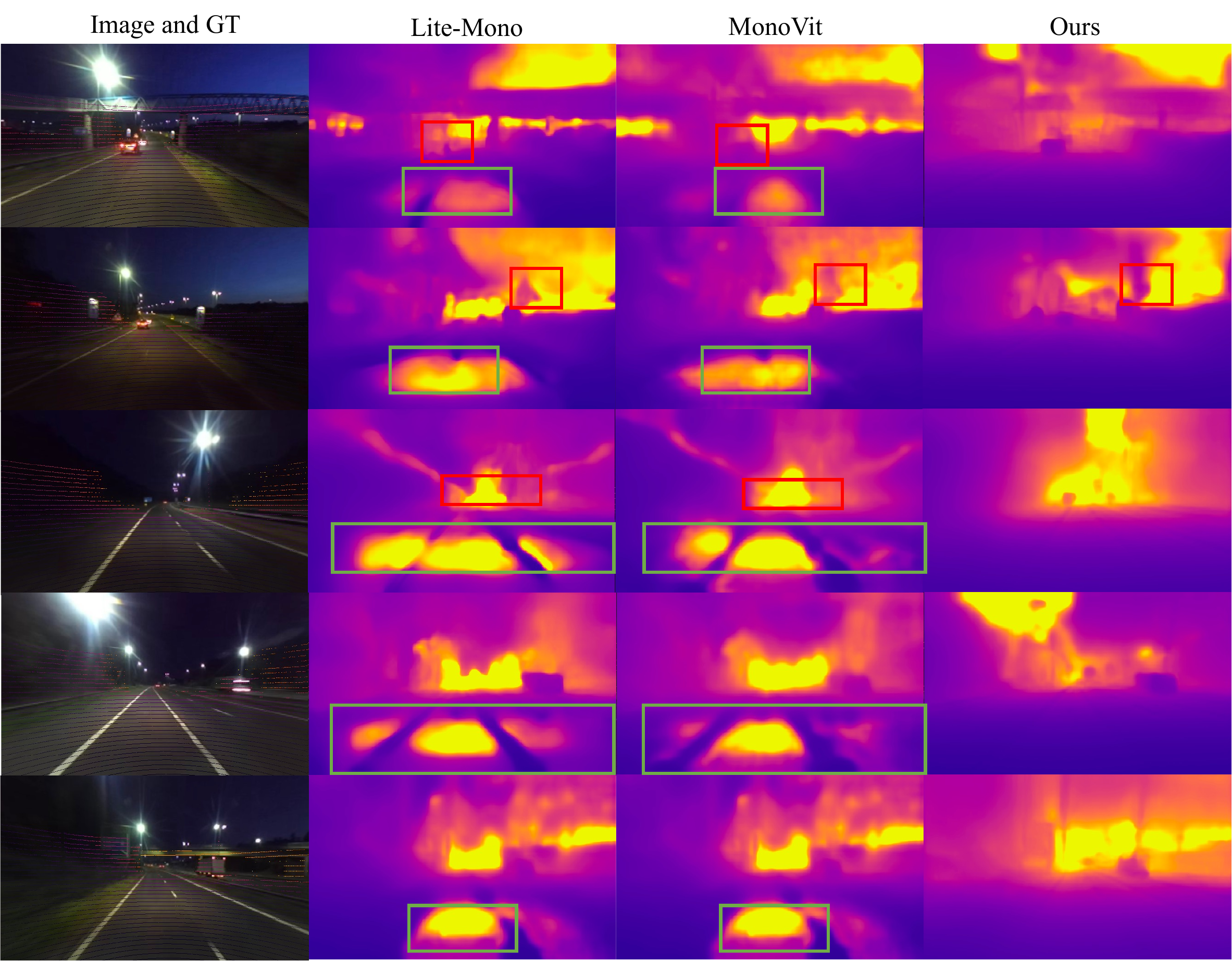}
    \caption{Visualization comparison under night conditions with different methods. Green boxes are used to highlight erroneous depth estimates, such as predicting the ground as a hole, while red boxes indicate ambiguous or missing object estimates, such as predicting the car and background as having the same depth. Our method provides stable predictions across all weather conditions.}
    \label{fig:night}
\end{figure*}

\begin{figure*}[htbp]
  \centering
   \includegraphics[width=0.6\linewidth]{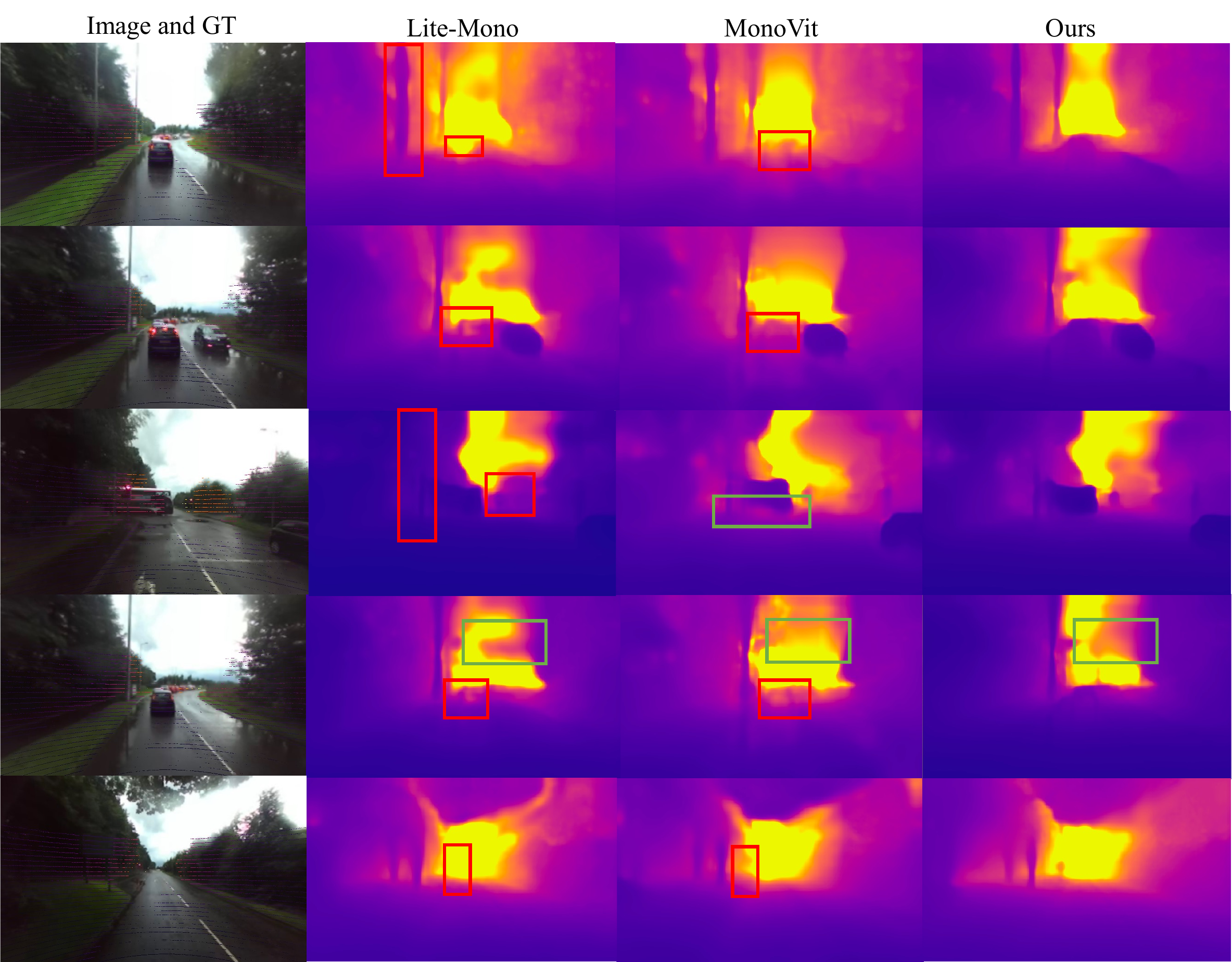}
    \caption{Visualization comparison under rainy conditions with different methods. Green boxes are used to highlight erroneous depth estimates, such as predicting the ground as a hole, while red boxes indicate ambiguous or missing object estimates, such as predicting the car and background as having the same depth. Our method provides stable predictions across all weather conditions.}
    \label{fig:rain}
\end{figure*}

\begin{figure*}[htbp]
  \centering
   \includegraphics[width=0.6\linewidth]{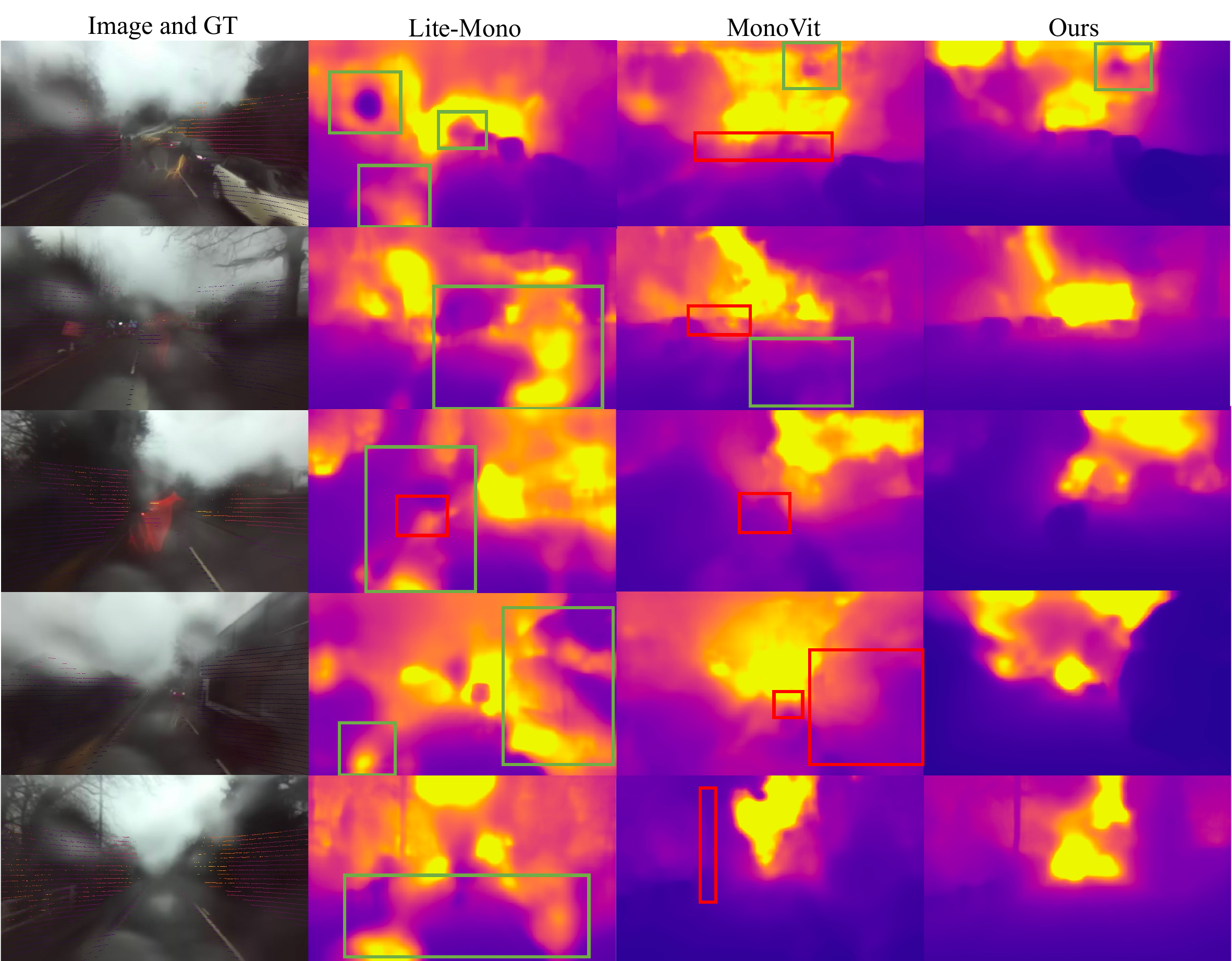}
    \caption{Visualization comparison under foggy conditions with different methods. Green boxes are used to highlight erroneous depth estimates, such as predicting the ground as a hole, while red boxes indicate ambiguous or missing object estimates, such as predicting the car and background as having the same depth. Our method provides stable predictions across all weather conditions.}
    \label{fig:fog}
\end{figure*}

\begin{figure*}[htbp]
  \centering
   \includegraphics[width=0.6\linewidth]{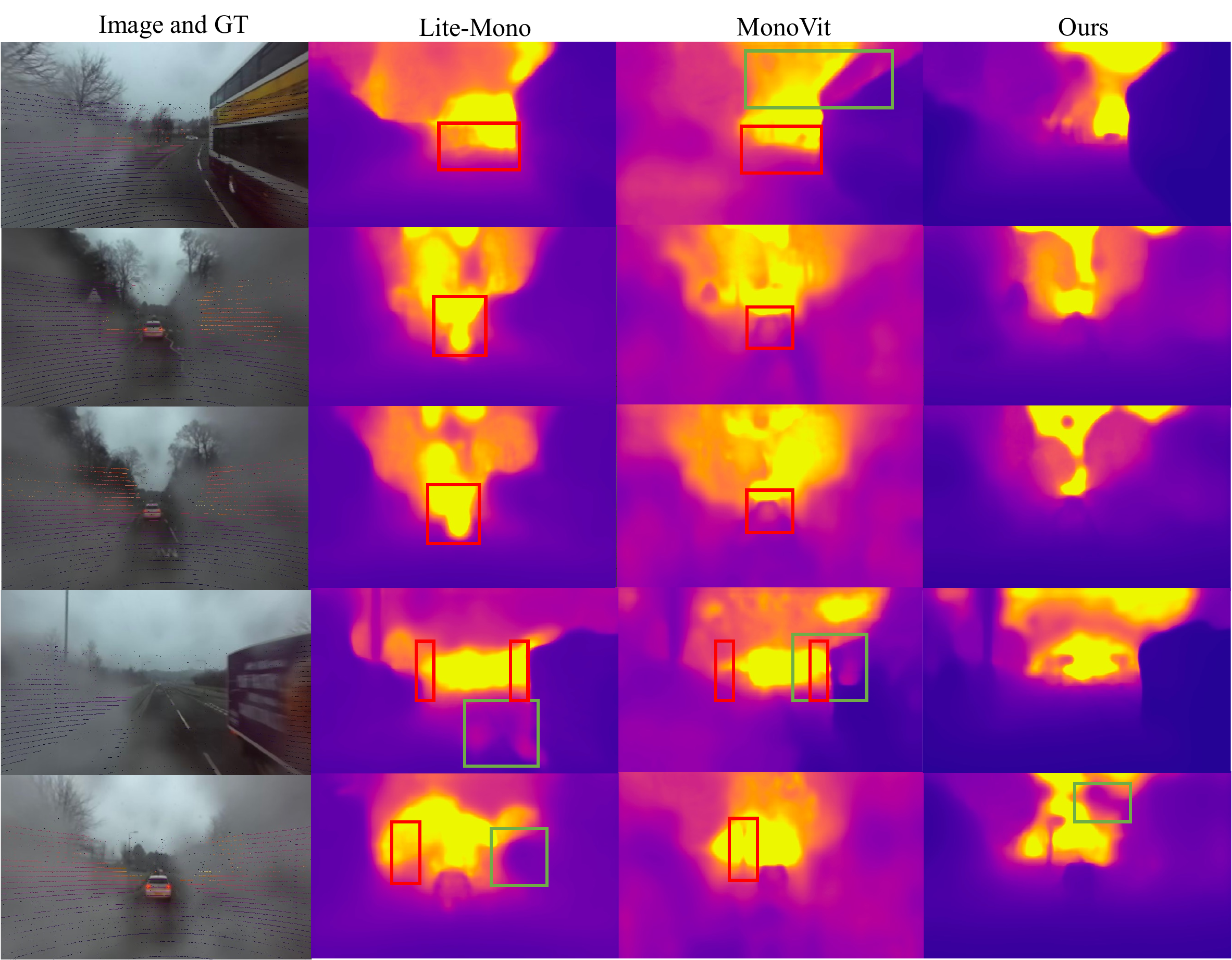}
    \caption{Visualization comparison under snowy conditions with different methods. Green boxes are used to highlight erroneous depth estimates, such as predicting the ground as a hole, while red boxes indicate ambiguous or missing object estimates, such as predicting the car and background as having the same depth. Our method provides stable predictions across all weather conditions.}
    \label{fig:snow}
\end{figure*}

\section{Additional Qualitative Results}
\label{sec:further_qual_comp}
In Fig.~\ref{fig:night}, Fig.~\ref{fig:rain}, Fig.~\ref{fig:fog}, and Fig.~\ref{fig:snow}, we provide additional visual comparisons under night, rain, fog, and snow conditions. We compare our proposed method applied to Lite-Mono with Lite-Mono and MonoVit, where all methods are trained on all-weather data. The wrong estimation caused by adverse conditions on depth prediction results is reflected in two aspects: first, models may incorrectly estimate regions such as raindrops or the ground at night, which we highlight with green boxes; second, models may produce ambiguous or even missing depth estimates for objects, such as vehicles under low visibility, which we mark with red boxes. Our method demonstrates stable predictions across all weather conditions.

\end{document}